\documentclass{article}
\usepackage[dvipsnames]{xcolor}
\usepackage{ijcai21}
\usepackage{times}

\usepackage{hyperref}
\usepackage{url}
\usepackage{graphicx}
\usepackage{outlines}
\usepackage{natbib}
\usepackage{booktabs}
\usepackage{times}
\usepackage{soul}
\usepackage{url}
\usepackage[utf8]{inputenc}
\usepackage[small]{caption}
\usepackage{graphicx}
\usepackage{amsmath}
\usepackage{amsthm}
\usepackage{booktabs}
\usepackage{subcaption}
\usepackage{float}
\usepackage[ruled,linesnumbered]{algorithm2e}
\usepackage{amsmath, amsthm, amssymb, amsfonts}

\usepackage{amsmath,amsfonts,bm}

\def\eqref#1{equation~\ref{#1}}
\def\Eqref#1{Equation~\ref{#1}}

\def\algref#1{algorithm~\ref{#1}}

\def\1{\bm{1}}

\newcommand{\thmref}[1]{Theorem~\ref{#1}}
\newcommand{\corref}[1]{Corollary~\ref{#1}}
\newcommand{\propref}[1]{Proposition~\ref{#1}}

\renewcommand{\algref}[1]{Algorithm~\ref{#1}}

\DeclareMathAlphabet{\mathsfit}{\encodingdefault}{\sfdefault}{m}{sl}
\SetMathAlphabet{\mathsfit}{bold}{\encodingdefault}{\sfdefault}{bx}{n}

\DeclareMathOperator*{\argmax}{arg\,max}
\DeclareMathOperator*{\argmin}{arg\,min}

\usepackage{etoolbox}

\newtheorem{theorem}{Theorem}
\newtheorem{corollary}[theorem]{Corollary}

\newtheorem{proposition}[theorem]{Proposition}

\theoremstyle{definition}

\theoremstyle{remark}

\numberwithin{equation}{section}

\renewcommand{\Pr}[1]{\ensuremath{\mathbb{P}\left[#1\right] }}

\def \argmax {\mathop{\rm arg\,max}}
\def \argmin {\mathop{\rm arg\,min}}

\newboolean{showcomments}
\setboolean{showcomments}{true}

\newcommand{\ziyu}[1]{\ifthenelse{\boolean{showcomments}}{\textcolor{purple}{[ZY: #1]}}{}}
\newcommand{\yuxin}[1]{\ifthenelse{\boolean{showcomments}}{\textcolor{blue}{[YC: #1]}}{}}

\newcommand{\cuparrow}{\color{ForestGreen}{{\boldsymbol{\uparrow}}}}
\newcommand{\cdownarrow}{\color{Maroon}{{\boldsymbol{\downarrow}}}}
\newcommand{\keep}{\color{White}{{\boldsymbol{\downarrow}}}}

\usepackage[dvipsnames]{xcolor}

\newcommand{\rebuttal}[1]{{\color{Maroon} #1}}
\newcommand{\ycnote}[1]{{\color{Maroon} #1}}
\newcommand{\yzy}[1]{{\color{blue} #1}}
\newcommand{\camera}[1]{{\color{Fuchsia} #1}}

\renewcommand{\rebuttal}[1]{#1}
\renewcommand{\ycnote}[1]{#1}
\renewcommand{\yzy}[1]{#1}
\renewcommand{\camera}[1]{#1}

\newcommand{\para}{\paragraph}

\newenvironment{packed_itemize}{
\begin{list}{\labelitemi}{\leftmargin=1.em}
  \setlength{\itemsep}{1pt}
  \setlength{\parskip}{0pt}
  \setlength{\parsep}{0pt}
  \setlength{\headsep}{0pt}
  \setlength{\topskip}{0pt}
  \setlength{\topmargin}{0pt}
  \setlength{\topsep}{0pt}
  \setlength{\partopsep}{0pt}
}{\end{list}}

\title{\rebuttal{Understanding the Effect of Bias in Deep Anomaly Detection}}

\author{
Ziyu Ye\footnote{Contact Author} \and
Yuxin Chen \And
Haitao Zheng \\
\affiliations
University of Chicago\\
\emails
\{ziyuye, chenyuxin\}@uchicago.edu,
htzheng@cs.uchicago.edu
}

\begin{document}
\maketitle


\begin{abstract}\vspace{-0.05in}

     Anomaly detection presents a unique challenge in machine learning, due to the scarcity of labeled anomaly data. Recent work attempts to mitigate such problems \rebuttal{by augmenting training of deep anomaly detection models with additional labeled anomaly samples}. However, the labeled data often does not align with the target distribution and introduces harmful bias to the trained model. In this paper, we aim to understand the effect of a biased anomaly set on anomaly detection. Concretely, we view 
     anomaly detection as a
     \rebuttal{supervised} learning task where the objective is to optimize the recall at a given false positive rate. 
     We formally study the \emph{relative scoring bias} of an anomaly detector, defined as the difference in performance with respect to a baseline anomaly detector.
     We establish the first finite sample rates for estimating the relative scoring bias for deep anomaly detection, and empirically validate our theoretical results on both synthetic and real-world datasets. We also provide an extensive empirical study on how a biased training anomaly set affects the anomaly score function and therefore the detection performance \rebuttal{on different anomaly classes.}
    Our study demonstrates scenarios in which the biased anomaly set can be useful \rebuttal{or problematic}, and provides a solid benchmark for future research. 
\vspace{-0.05in}
\end{abstract}

\section{Introduction}

Anomaly detection\rebuttal{~\citep{anormaly-survey-2009}} trains a formal model to identify unexpected or anomalous  instances in incoming data, whose behavior differs from normal instances. It is particularly useful for detecting problematic events such as digital fraud, structural defects, and system malfunctions.  Building accurate anomaly detection models is a well-known challenge in machine learning, due to the scarcity of labeled anomaly data.  The classical and most common approach is to train anomaly detection models using \rebuttal{only normal data}\footnote{\rebuttal{Existing literature has used different terms to describe such models, e.g., semi-supervised anomaly detection~\citep{anormaly-survey-2009} and unsupervised anomaly detection~\citep{pmlr-v80-ruff18a}.}}, i.e., first train a model using a corpus of normal data to capture {\em normal} behaviors, then configure the model to flag instances with large deviations as anomalies.  Researchers have also developed deep learning methods to better capture the complex structure in the data \citep{pmlr-v80-ruff18a, Zhou2017AE}.  \rebuttal{Following the terminology introduced by~\cite{anormaly-survey-2009}, we refer to these models as deep \emph{semi-supervised} anomaly detection models.}

Recently, a new line of anomaly detection models \rebuttal{propose to leverage available labeled anomalies during model training, i.e., train an anomaly detection model using both normal data and additional labeled anomaly samples} as they become available~\citep{ruff2020deep,yamanaka2019autoencoding,ruff2020rethinking}. Existing works show that these new models achieve considerable performance improvements beyond the models trained using only normal data.  \rebuttal{We hereby refer to these models as deep  \emph{supervised} anomaly detection  models\footnote{\rebuttal{Some works termed these models as semi-supervised anomaly detection~\citep{ruff2020deep,yamanaka2019autoencoding,ruff2020rethinking} while others termed them as supervised anomaly detection~\citep{anormaly-survey-2009}.}}~\citep{anormaly-survey-2009}.}

When exploring these models, we found when the labeled anomalies (used to train the model) do not align with the target distribution \yzy{(typically unknown)}, they can introduce harmful bias to the trained model. Specifically, when comparing the performance of a supervised anomaly detector to its \rebuttal{semi-supervised version},  the performance difference varies significantly across test anomaly data, some better and some worse.  That is, \rebuttal{using labeled anomalies during model training does not always improve model performance}; instead, it may introduce unexpected bias in anomaly detection outcomes.

\begin{table*}[tp]
\centering
\vspace{-0.1in}
\resizebox{\textwidth}{!}{%
\begin{tabular}{@{}cccc@{}}
\toprule
\textbf{Task Type} & \textbf{Distribution Shift} & \textbf{Known Target Distribution} & \textbf{Known Target Label Set} \\ \midrule
{\bf Imbalanced Classification}~\citep{johnson2019survey} & No & N/A & N/A \\
{\bf Closed Set Domain Adaptation}~\citep{saenko2010adapting} & Yes & Yes & Yes \\
{\bf Open Set Domain Adaptation} ~\citep{panareda2017open}         & Yes & Yes & No \\
{\bf Anomaly Detection}~\citep{chalapathy2019deep}             & Yes & No & No \\ \bottomrule
\end{tabular}%
}
\caption{Comparison of anomaly detection tasks with other relevant classification tasks.}
\label{tab:unique-anomaly}
\vspace{-0.2in}
\end{table*}

In this paper, we aim to devise a \textit{rigorous} and \textit{systematic} understanding on the effect of \yzy{labeled anomalies} on deep anomaly detection models. We formally state the anomaly detection problem as a learning task aiming to optimize the recall of anomalous instances at a given false positive rate---a performance metric commonly used by many real-world anomaly detection tasks~\citep{liu2018open,limobihoc2019}. 
We then show that different types of anomalous labels produce different anomaly scoring functions. Next, given \emph{any} reference anomaly scoring function, we formally define the \emph{relative scoring bias} of an anomaly detector as its difference in performance with the reference scoring function. 
%

Following our definition\footnote{Our definition of scoring bias for anomaly detection aligns with the classical notion of bias in the supervised learning setting, with the key difference being the different performance metric.},
we establish the first finite sample rates for estimating the relative scoring bias for deep anomaly detection. We empirically validate our assumptions and theoretical results on both synthetic and three real-world datasets \rebuttal{(Fashion-MNIST, StatLog~(Landsat Satellite), and Cellular Spectrum Misuse~\citep{limobihoc2019}).}

Furthermore, we provide an extensive empirical study on how additional labeled data affects the anomaly score function and the resulting detection performance.  We consider the above three real-world datasets and six deep anomaly detection models. Our study demonstrates a few typical scenarios in which the \yzy{labeled anomalies} can be useful \rebuttal{or problematic}, and provides a solid benchmark for future research.
%
Our main contributions are as follows:
\begin{packed_itemize} \vspace{-0.06in}
    \item We systematically expose the bias effect and discover the issue of large performance variance in deep anomaly detectors, caused by the use of the \yzy{additional labeled anomalies} in training.

    \item We model the effect of biased training as relative scoring bias, and establish the first finite sample rates for estimating the relative scoring bias of the trained models.

    \item We conduct empirical experiments to verify and characterize the impact of the relative scoring bias on six popular anomaly detection models, and three real-world datasets.  \vspace{-0.06in}
\end{packed_itemize} 

\rebuttal{To the best of our knowledge, we are the first to formally study the effect of additional labeled anomalies on deep anomaly detection.
Our results show both significant positive and negative impacts from them, and suggest that model trainers must treat additional labeled data with extra care. We believe this leads to new opportunities for improving deep anomaly detectors and deserves more attention from the research community. }

\section{Related Work}
\para{\bf Anomaly detection models.} \rebuttal{While the literature on anomaly detection models is extensive, the most relevant to our work are deep learning based models.
Following the terminology used by~\cite{anormaly-survey-2009}, we consider two types of models: \vspace{-0.06in}
\begin{packed_itemize}
\item \camera{{\em Semi-supervised anomaly detection} refers to models trained on only normal data, e.g., ~\cite{Zhou2017AE, pmlr-v80-ruff18a, goyal2020drocc}};
\item \camera{{\em Supervised anomaly detection} refers to models trained on  normal data and a small set of labeled anomalies~\citep{pang2019deep, yamanaka2019autoencoding, ruff2020rethinking,ruff2020deep, goyal2020drocc}}. 
Due to the increasing need of making use of labled anomalies in real-world applications, this type of work has gain much attention recently.
\vspace{-0.06in}
\end{packed_itemize}
}

Another line of recent work proposes to use synthetic anomalies \citep{golan2018deep, hendrycks2019using, lee2018training}, ``forcing'' the model to learn a more compact representation for normality. While the existing work has shown empirically additional labeled anomalies in training may help detection, it does not offer any theoretical explanation, nor does it consider the counter-cases when additional labeled anomalies hurt detection.

Deep anomaly detection models can also be categorized by architectures and objectives, e.g., hypersphere-based models \citep{pmlr-v80-ruff18a,ruff2020rethinking,ruff2020deep} and autoencoder (or reconstruction) based models~\citep{Zhou2017AE,yamanaka2019autoencoding} (see Table~\ref{tab:modeldata}). We consider both types in this work.


\para{\bf Bias in anomaly detection.} While the issue of domain mismatch has been extensively studied as transfer learning in general supervised learning scenarios, it remains an open challenge for anomaly detection tasks.
\camera{Existing work on anomaly detection has explored bias in {\em semi-supervised} setting when noise exists in normal training data \citep{tong2019fixing, liu2019exploring}, but little or no work has been done on the {\em supervised} setting \rebuttal{(i.e., models trained on both normal data and some labeled anomalies)}. Other well-studied supervised tasks, as summarized in Table~\ref{tab:unique-anomaly}, generally 
assume that one can draw representative samples from the target domain.} 
Unlike those studies, anomaly detection tasks are constrained by limited information on unknown types of anomalies in testing, thus additional labeled data in training can bring significant {\em undesired} bias. This poses a unique challenge in inferring and tackling the impact of bias in anomaly detection (e.g., defending against potential data poisoning attacks). To the best of our knowledge, we are the first to identify and systematically study the bias caused by an additional (possibly unrepresentative) labeled anomaly set in deep anomaly detection models (as shown in Section~\ref{sec:eval}).




\para{\bf PAC guarantees for anomaly detection.} Despite significant progress on developing theoretical guarantees for classification \citep{valiant1984theory}, little has been done for anomaly detection tasks. \cite{siddiqui2016finite} first establish a PAC framework for anomaly detection models by the notion of pattern space; however, \camera{it is challenging to be generalized to deep learning frameworks.} \cite{liu2018open} propose a model-agnostic approach to provide PAC guarantees for unsupervised anomaly detection performance. 
\camera{We follow the basic setting from this line of work to address the convergence of the relative scoring bias. Closely aligned with the empirical risk minimization framework \citep{vapnik1992principles}, our definition for bias facilitates connections to fundamental concepts in learning theory and brings rigor in theoretical study of anomaly detection.} In contrast to prior work, our proof relies on a novel adaption of the key theoretical tool from \cite{massart1990}, which allows us to extend our theory to characterize the notion of scoring bias as defined in Section~\ref{sec:defbias}.

%
%

\section{Problem Formulation}

\newcommand{\loss}{\ell}
\newcommand{\trainingset}{D}
\newcommand{\Models}{\ensuremath{\Theta}}
\newcommand{\model}{\theta}
\newcommand{\instance}{x}
\newcommand{\clabel}{y}
\newcommand{\score}{s}
\newcommand{\threshold}{\tau}
\newcommand{\FPR}{\text{FPR}\xspace}
\newcommand{\TPR}{\text{TPR}\xspace}
\newcommand{\Recall}{\text{Recall}\xspace}
\newcommand{\quota}{q}
\newcommand{\TrueDist}{\mathcal{D}}
\newcommand{\DataDist}{\mathcal{D}}
\newcommand{\Dataset}{D}
\newcommand{\val}{\text{val}}
\newcommand{\bias}{\text{bias}}
\newcommand{\iid}{\textit{i.i.d.}\xspace}

\ycnote{We now formally state the anomaly detection problem}. Consider a model class $\Models$ for anomaly detection, and a (labeled) training set $\trainingset$ sampled from a mixture distribution $\DataDist$ over the normal and anomalous instances. A model ${\model}$ maps each input instance $\instance$ to a continuous output, which corresponds to anomaly score $\score_{\model}(\instance)$. The model uses a threshold $\threshold_{\model}$ on the score function to produce a binary label for $\instance$.

Given a threshold $\threshold_\model$, we define the False Positive Rate (FPR) of $\model$ on the input data distribution as
$\FPR(\score_\model, \threshold_\model) = \Pr{\score_\model(x) > \threshold_\model \mid y = 0 }$, and the True Positive Rate (TPR, a.k.a. Recall) as
$\TPR(\score_\model, \threshold_\model) = \Pr{\score_\model(x) > \threshold_\model \mid y = 1 }$.
The FPR and TPR are competing objectives---thus, a key challenge for anomaly detection algorithms is to identify a configuration of the score, threshold pair $(\score_\model, \threshold_\model)$ that strikes a balance between the two metrics.
W.l.o.g.\footnote{Our results can be easily extended to the setting where the goal is to minimize \FPR subject to a given \TPR~\camera{(cf. Appendix~\ref{sec:proofthm}}).}, in this paper we focus on the following scenario, where the objective is to maximize \TPR subject to a target $\FPR$. Formally, let $1 - \quota$ be the target \FPR; we define the optimal anomaly detector as\footnote{\ycnote{This formulation aligns with many contemporary works in deep anomaly detection. For example, \citet{limobihoc2019} show that in real world, it is desirable to detect anomalies with a prefixed low false alarm rate; 
\citet{liu2018open} formulate anomaly detection in a similar way, where the goal is to minimize FPR for a fixed TPR.}}
\begin{align}
    (\score^*_\model,\threshold^*_\model) \in \argmax_{(\score_\model,\threshold_\model): \model \in \Models} \TPR(\score_\model,\threshold_\model) \text{~s.t.~} \FPR(\score_\model, \threshold_\model) \leq 1 - q. \label{eq:objective}
\end{align}

\subsection{A General Anomaly Detection Framework}\label{sec:twostage}
The performance metric (namely \TPR) in Problem~\ref{eq:objective} depends on the entire predictive distribution, and can not be easily evaluated on any single data point. Thus, rather than directly solving Problem~\ref{eq:objective}, practical anomaly detection algorithms (e.g., Deep SVDD \citep{pmlr-v80-ruff18a}) often rely on a two-stage process: (1) learning the score function $\score_\model$ from training data via a surrogate loss, and (2) given $\score_\model$ from the previous step, computing the threshold function $\threshold_\model$ on the training data. Formally, given a model class $\Models$, a training set $\trainingset$, a loss function $\loss$, and a target \FPR $1 - \quota$, a two-staged anomaly detection algorithm outputs:
\begin{align}
\begin{cases}
\hat{\score}_\model \in \argmin_{\score_\model: \model\in \Models} \loss(\score_\model,\trainingset) \\
\hat\threshold_\model \in \argmax_{\threshold_\model: \model \in \Models} \TPR(\hat\score_\model,\threshold_\model) \text{~s.t.~} \FPR(\hat\score_\model, \threshold_\model) \leq 1 - q. \label{eq:twostage} 
\end{cases}
\end{align}
\ycnote{The first part of \Eqref{eq:twostage} amounts to solving a supervised learning problem.} Here, the loss function $\loss$ could be instantiated into latent-space-based losses (e.g., Deep SVDD), margin-based losses (e.g., OCSVM~\citep{ocsvm}), or reconstruction-based losses (e.g., ABC~\citep{yamanaka2019autoencoding});
therefore, many contemporary anomaly detection models fall into this framework. To set the threshold $\hat\threshold_\model$, we consider using the distribution of the anomaly scores $\hat\score_\model(\cdot)$ from a labeled validation set $\Dataset^{\val} \sim \TrueDist$. Let $\Dataset^{\val} := \Dataset^{\val}_{0}\cup \Dataset^{\val}_{a} $ where $\Dataset^{\val}_{0}$ and $\Dataset^{\val}_{a}$ denote the subset of normal data and the subset of abnormal data of $\Dataset^{\val}$. 
Denote the empirical CDFs for anomaly scores assigned to $x$ in $\Dataset^{\val}_{0}$ and $\Dataset^{\val}_{a}$ 
as $\hat{F}_{0}$ and $\hat{F}_{a}$,  
respectively. Given a target \FPR value $1 - \quota$, \camera{similar to} \cite{liu2018open}, one can compute the threshold as $\hat{\tau}_\model=\max \{ u \in \mathbb{R}:\hat{F}_{0}(u) \leq q \}$. 
\algref{alg:twostage-part2} summarizes the steps to solve the second part of \Eqref{eq:twostage}.
\vspace{-0.06in}
\begin{algorithm}[h!]
  \SetAlgoLined
  \KwData{A validation dataset $\Dataset^{\val}$ and a scoring function $s(\cdot)$.}
  \KwResult{A score threshold achieving a target FPR and the corresponding recall on $\Dataset^{\val}$.}
  \BlankLine
  Get anomaly score $s(x)$ for each x in $\Dataset^{\val}$.

  Compute empirical CDF $\hat{F}_{0} (x)$ and $\hat{F}_{a} (x)$ for anomaly scores of $x$ in $\Dataset^{\val}_{0}$ and $\Dataset^{\val}_{a}$.

  Output detection threshold $\hat{\tau}=\max \{ u \in \mathbb{R}:\hat{F}_{0}(u) \leq q \}$.

  Output \ycnote{\TPR} (recall) 
  on $\Dataset^{\val}_{a}$ as $\hat{r} = 1 -  \hat{F}_{a}(\hat{\tau})$.

  \caption{
  Computing the anomaly detection threshold for Problem~\ref{eq:twostage}
  }\label{alg:twostage-part2}
\end{algorithm}\vspace{-0.06in}


\subsection{Scoring Bias}\label{sec:defbias}
\camera{Given a model class $\Models$ and a training set $\trainingset$, 
we define the \emph{scoring bias} of a detector $(\hat\score_\model, \hat\threshold_\model)$ as:}
\begin{align}\label{eq:bias}
    \bias(\hat\score_\model, \hat\threshold_\model) := \argmax_{(\score_\model,\threshold_\model): \model \in \Models} \TPR(\score_\model,\threshold_\model) - \TPR(\hat\score_\model,\hat\threshold_\model).
\end{align}
We call $(\hat\score_\model, \hat\threshold_\model)$ a \emph{biased} detector if $\bias(\hat\score_\model, \hat\threshold_\model) > 0$. In practice, due to the biased training distribution and the fact that the two-stage process in \ycnote{\Eqref{eq:twostage}} is not directly optimizing \TPR, the resulting anomaly detectors are often biased by construction. One practically relevant performance measure is the \emph{relative scoring bias}, defined as the difference in \TPR between two anomaly detectors, subject to the constraints in \Eqref{eq:twostage}. It captures the relative strength of two algorithms in detecting anomalies, and thus is an important indicator for model evaluation and selection\footnote{\camera{This TPR-based definition for bias is also useful for group fairness study. For example, in Figure~\ref{fig:scenario}, we have shown how model performances vary across different \textit{sub-groups} of anomalies.}}. Formally, given two \ycnote{\emph{arbitrary}} anomaly score functions $\score, \score'$ and corresponding threshold functions $\threshold, \threshold'$ obtained from \algref{alg:twostage-part2}, we define the \emph{relative scoring bias} between $\score$ and $\score'$ as:
\begin{align}\label{eq:relativebias}
    \xi(\score, \score') &:= \bias(\score, \threshold) - \bias(\score', \threshold') \nonumber \\
    &= \TPR(\score',\threshold') - \TPR(\score,\threshold).
\end{align}
Note that when $\score' = \score^*_\model$, the relative scoring bias (\eqref{eq:relativebias}) reduces to the scoring bias (\eqref{eq:bias}). We further define the \emph{empirical relative scoring bias} between $\score$ and $\score'$ as
\begin{align}\label{eq:emprelativebias}
    \hat{\xi}(\score, \score') := \widehat{\TPR}(\score',\threshold') - \widehat{\TPR}(\score,\threshold),
\end{align}
where $\widehat\TPR(\score,\threshold)=\frac{1}{n} \sum_{j=1}^{n} \mathbf{1}_{\ycnote{\score(x_{j}) > \threshold};y_{j}=1}$ 
denotes the \TPR (recall) estimated on a finite validation set of size $n$. In the following sections, we will investigate both the theoretical properties and the empirical behavior of the empirical relative scoring bias for contemporary anomaly detectors.

\begin{table*}[t!]
\centering
\vspace{-0.1in}
\resizebox{\textwidth}{!}{
\begin{tabular}{c|c|c}
\toprule
      {\bf Type}              & \rebuttal{{\bf Semi-supervised (trained on normal data)}}   &  \rebuttal{{\bf Supervised (trained on normal \& some abnormal data)}}                                                                                        \\ \hline
Hypersphere-based    & Deep SVDD~\citep{pmlr-v80-ruff18a}        & {\begin{tabular}[c]{@{}c@{}} Deep SAD~\citep{ruff2020deep}, Hypersphere Classifier (HSC)~\citep{ruff2020rethinking} \end{tabular}}             \\ \hline
Reconstruction-based & Autoencoder (AE)~\citep{Zhou2017AE} & {\begin{tabular}[c]{@{}c@{}} Supervised AE (SAE)\footnote{We design SAE by forcing the reconstruction errors to be maximized for additional labeled anomalies encountered in training the autoencoder.}, Autoencoding Binary Classifier (ABC)\end{tabular}}~\citep{yamanaka2019autoencoding} \\ \toprule
\end{tabular}
}
\vspace{-0.1in}
\caption{The anomaly detection models considered in our case study. Deep SAD and HSC are the \ycnote{supervised versions} of Deep SVDD (the \ycnote{semi-supervised}  baseline model); SAE and ABC are the  \ycnote{supervised versions} of AE (the \ycnote{semi-supervised} baseline model).}
\label{tab:modeldata}
\vspace{-0.1in}
\end{table*}

\section{Finite Sample Analysis for Empirical Relative Scoring Bias}
\label{sec:pac}

In this section, we show how to estimate the relative scoring bias (\Eqref{eq:relativebias}) given \ycnote{\emph{any}} two scoring functions $s,s'$. As an example, $s$ could be \ycnote{induced by a semi-supervised anomaly detector trained on normal data only}, and $s'$ could be \ycnote{induced by a supervised anomaly detector trained on a biased anomaly set}. We then provide a finite sample analysis of the convergence rate of the empirical relative scoring bias, and validate our theoretical analysis via a case study.

\subsection{Finite Sample Guarantee}
\label{subsec:finite}
\para{\bf Notations.} Assuming when determining $\hat{\tau}$, scoring functions $s,s'$ are evaluated on the unbiased empirical distribution of normal data; the empirical \TPR $\hat{r}$ are estimated on the unbiased empirical distribution of abnormal data.
Let $\{ s_i := s(x_i) \mid x_i, y_{i}=0 \}_{i=1} ^{n_0}$ be anomaly scores evaluated by $s(\cdot)$ on $n_0$ \iid random normal data points. Let $F_{0}(t) := \Pr{s(x) \ycnote{\leq} t\mid y=0}$ be the CDF of $s(x)$, and $\hat{F}_{0}(t) :=\frac{1}{n_0} \sum_{i=1}^{n_0} \mathbf{1}_{s_{i} \leq t;y_{i}=0}$ be its empirical CDF.
For $n_1$ \iid samples $\{s_{j} := s(x_j) \mid x_j, y_j=1 \} _{j=0} ^{\ycnote{n_1}}$, we denote the CDF as $F_{a}(t) := \Pr{s(x) \ycnote{\leq} t\mid y=1}$,  and the emprical CDF as $\hat{F}_{a}(t) :=\frac{1}{n_1} \sum_{j=1}^{n_1} \mathbf{1}_{s_{j} \leq t;y_{j}=1}$. Similarly, we denote the CDF and emiprical CDF for $\{s'_{i} \mid y_{i}=0 \}_{i=0} ^{n_0}$ as ${F}'_{0}(t)$ and $\hat{{F}}'_{0}(t)$, and for $\{ {s'}_{j} \mid y_{j}=1 \}_{j=0} ^{n_1}$ as $F'_{a}(t)$ and $\hat{F}'_{a}(t)$, respectively.

\para{\bf Infinite sample case.} In the limit of infinite data (both normal and abnormal), $\hat{{F}}_{0}, \hat{{F}}_{a}, \hat{F}'_{0}, \hat{F}'_{a}$ converges to the true CDFs (\ycnote{cf. Skorokhod's representation theorem and Theorem 2A of \citet{parzen1980quantile}}), 
and hence the empirical relative scoring bias also converges. \propref{prop:infinite} establishes a connection between the CDFs and the relative scoring bias.
\begin{proposition}\label{prop:infinite}
Given two scoring functions $s, s'$ and a target \FPR $1 - \quota$, the relative scoring bias \ycnote{is} $\xi(\score, \score') = F_{a}(F_{0}^{-1}(q)) - F'_{a}({{F'_{0}}^{-1}(q)})$.
\end{proposition}
Here, $F^{-1}(\cdot)$ is the quantile function. The proof of \propref{prop:infinite} follows from the fact that for corresponding choice of $\threshold, \threshold'$ in \algref{alg:twostage-part2}, $\TPR(\score, \threshold)=1-F_{a}(F_{0}^{-1}(q))$, and $\TPR\ycnote{(\score', \threshold')}=1-F'_{a}({{F'_{0}}^{-1}(q)})$.

Next, a direct corollary of the above result shows that,  for the special cases where both the scores for normal and abnormal data are Gaussian distributed,  one can directly compute the relative scoring bias. The proof is listed in Appendix~\ref{sec:proof-cor2}.
\begin{corollary}\label{cor:infinite-gaussian}
\ycnote{Let $1 - \quota$ be a fixed target \FPR.}  Given two scoring functions $s, s'$, assume that $s(x) \mid (y=0) \sim \mathcal{N}(\mu_0, \sigma_0)$, $s(x) \mid (y=1) \sim \mathcal{N}(\mu_a, \sigma_a)$, $s'(x) \mid (y=0) \sim \mathcal{N}(\mu'_0, \sigma'_0)$, $s'(x) \mid (y=1) \sim \mathcal{N}(\mu'_a, \sigma'_a)$.
\ycnote{Then, } the relative scoring bias is
\begin{equation*}\resizebox{.96\hsize}{!}{$\xi(\score, \score') = \Phi\left(\frac{\sigma_{0}\Phi ^{-1}(q)}{\sigma_{a}} + \frac{\mu_{0} - \mu_{a}}{\sigma_{a}}\right) -\Phi\left(\frac{{\sigma'}_{0}\Phi ^{-1}(q)}{{\sigma'}_{a}} + \frac{{\mu'}_{0} - {\mu'}_{a}}{{\sigma'}_{a}}\right)$},\end{equation*}
\ycnote{where $\Phi$ denotes the CDF of the standard Gaussian.}
\end{corollary}


\para{\bf Finite sample case.} In practice, when comparing the performance of two scoring functions, we only have access to finite samples. Thus, it is crucial to bound the estimation error due to insufficient samples. We now establish a finite sample guarantee for estimating the relative scoring bias. Our result extends the analysis of \cite{liu2018open}. The validation set contains a mixture of $n=n_0 + n_1$ i.i.d. samples, with $n_0$ normal samples and $n_1$ abnormal samples where $\frac{n_1}{n} = \alpha$.

 The following result shows that under mild assumptions of the  continuity of the CDFs and quantile functions, the sample complexity for achieving ${|\hat\xi - \xi| \leq \epsilon}$:
\begin{theorem}\label{thm:scomplx}
Assume that $F_a, F'_a, F_0^{-\ycnote{1}}, {F'_0}^{-\ycnote{1}}$ are Lipschitz continuous with Lipschitz constant $\ell_a, \ell'_a, \ell_0^-, {\ell'_0}^-$, respectively. Let $\alpha$ be the fraction of abnormal data among $n$ \iid samples from the mixture distribution. Then, w.p. at least $1-\delta$, with
\begin{equation*}\resizebox{.96\hsize}{!}{
    $n \geq \frac{8}{\epsilon^2} \cdot
    \left( \log \frac{2}{1-\sqrt{1-\delta}} \cdot \left(\frac{2-\alpha}{\alpha} \right)^2 + \log \frac{2}{\delta} \cdot \frac{1}{1-\alpha} \left(\left( \frac{\ell_a}{\ell_0^-} \right)^2 + \left( \frac{\ell'_a}{{\ell'_0}^-} \right)^2\right) \right)$},
\end{equation*}


the empirical relative scoring bias satisfies $|\hat\xi - \xi| \leq \epsilon$.
\end{theorem}
We defer the proof of \thmref{thm:scomplx} to Appendix~\ref{sec:proofthm}. The sample complexity for estimating the relative scoring bias $n$ grows as $\mathcal{O}\left( \frac{1}{\alpha^2\epsilon^2} \log\frac{1}{\delta}\right)$. Note the analysis of our bound involves a novel two-step process which first bounds the estimation of the threshold for the given \FPR, and then leverages the Lipschitz continuity condition to derive the final bound.

\subsection{Case Study}
\label{sec:analysis-casestudy}
We conduct a case study to validate our main results above using a synthetic dataset and \rebuttal{three real-world datasets.}
We consider six anomaly detection models listed in Table~\ref{tab:modeldata}, and they lead to consistent results.  \rebuttal{For brevity, we show results when using Deep SVDD as the baseline model trained on normal data only and Deep SAD as the supervised model trained on normal and some abnormal data.} Later in  Appendix~\ref{section:pac-additional}, we include results of other models, including Deep SVDD vs. HSC, AE vs. SAE, and AE vs. ABC.

\para{\bf Synthetic dataset.} Similar to~\cite{liu2018open}, we generate our synthetic dataset by sampling data from a mixture data distribution $S$, w.p. $1-\alpha$ generating the normal data distribution $S_{0}$ and w.p. $\alpha$ generating the abnormal data distribution $S_{a}$. Data in $S_{0}$ are sampled randomly from a 9-dimensional Gaussian distribution, where each dimension is independently distributed as $\mathcal{N}(0, 1)$. Data in $S_{a}$ are sampled from another 9-dimensional distribution, which w.p. 0.4 have 3 dimensions (uniformly chosen at random) distributed as $\mathcal{N}(1.6, 0.8)$, w.p. 0.6 have 4 dimensions (uniformly chosen at random) distributed as $\mathcal{N}(1.6, 0.8)$, and have the remaining dimensions distributed as $\mathcal{N}(0, 1)$. This ensures meaningful feature relevance, point difficulty and variation for the abnormal data distribution~\citep{emmott2015meta}.

We obtain score functions $s$ and $s'$ by training Deep SVDD and Deep SAD respectively on samples from the synthetic dataset (10K data from $S_{0}$, 1K data from $S_{a}$). We configure the training, validation and test set so they have no overlap. Thus the training procedure will not affect the sample complexity for estimating the bias. To set the threshold, we fix the target FPR to be 0.05,  and vary the number of normal data in the validation set $n$ from \{100, 1K, 10K\}. We then test the score function and threshold on a fixed test dataset with a large number (20K) of normal data and $\alpha\times$20K of abnormal data. We vary $\alpha$ from \{0.01, 0.05, 0.1, 0.2\}.

\para{\bf Real-world datasets.} We consider 3 real-world datasets targeting disjoint subjects: Fashion-MNIST~\citep{xiao2017fashionmnist}, StatLog~\citep{Statlogdataset} and Cellular Spectrum Misuse~\citep{limobihoc2019}. Detailed descriptions of datasets and training configurations are in Appendix~\ref{sec:datasets}. 

\begin{figure}[ht]
\centering
 \includegraphics[width=0.45\textwidth]{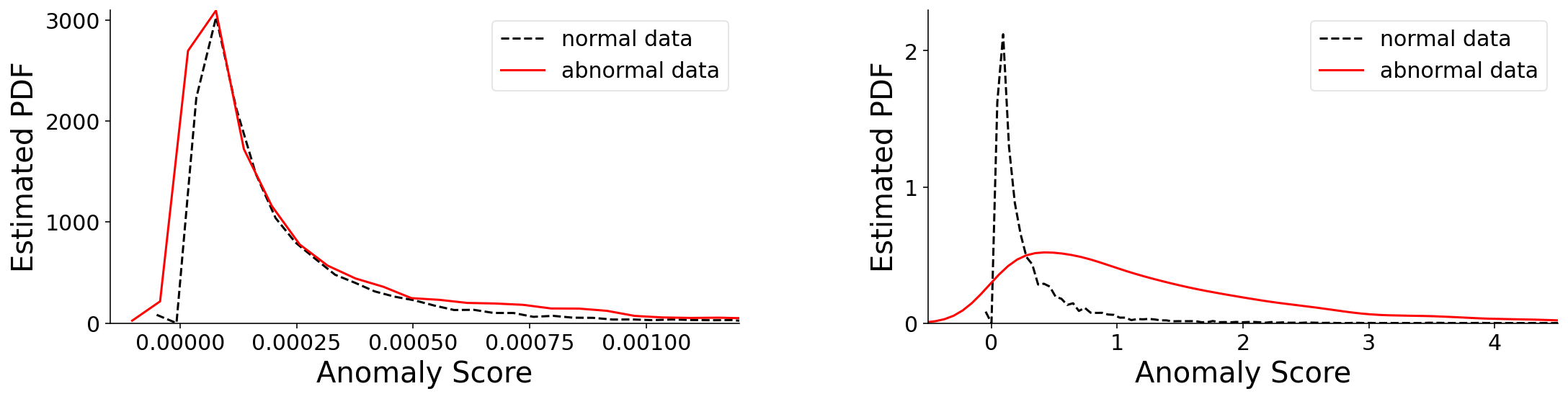}
 \vspace*{-2mm}
\caption{Anomaly score distributions for Deep SVDD (left) and Deep SAD (right) on the synthetic dataset.}
\label{fig:pdf}
\end{figure}

\para{\bf Distribution of anomaly scores.} Figure~\ref{fig:pdf} is a sample plot of score distributions on the test set of the synthetic dataset with $\alpha=0.1$. We make two key observations. First, the distribution curves follow a rough bell shape. Second and more importantly, while the abnormal score distribution can closely mimic the normal score distribution under the unsupervised model, it deviates largely from the normal score distribution after semi-supervised training.  This confirms that semi-supervised training does introduce additional bias.

\vspace{-0.10in}


We also examine the anomaly score distributions for models trained \ycnote{on real-world datasets, including Fashion-MNIST and Cellular Spectrum Misuse}. While the score distributions are less close to Gaussian, we do observe the same trend where normal and abnormal score distributions become significantly different after applying semi-supervised learning. The results are shown in Figure~\ref{fig:pdf-fmnist} and \ref{fig:pdf-spectrum} in Appendix~\ref{section:pac-additional}.



\begin{figure}[h]
\centering
 \includegraphics[width=0.45\textwidth]{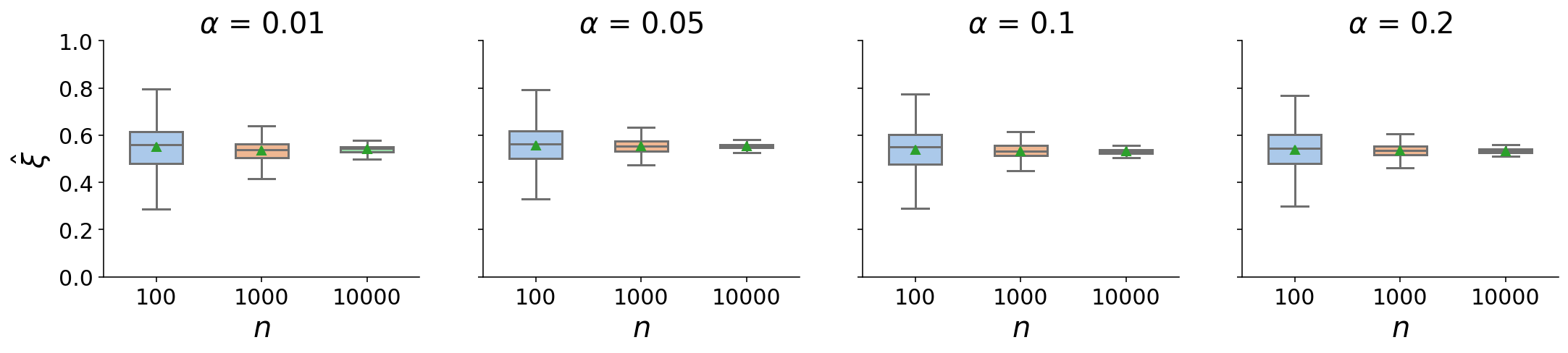}
 \includegraphics[width=0.45\textwidth]{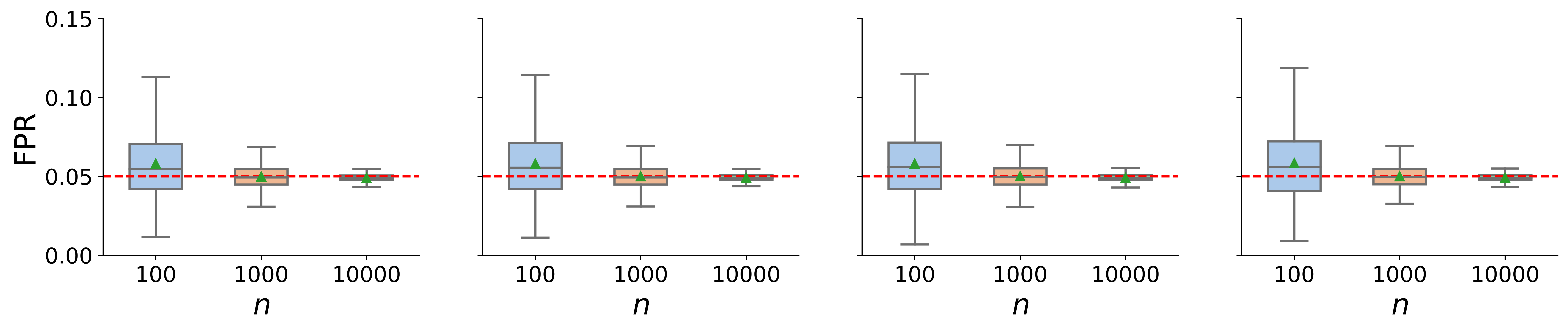}
\caption{Models \rebuttal{(Deep SVDD v.s. Deep SAD)} trained on the synthetic dataset:  the quantile distribution of  relative scoring bias $\hat{\xi}$ (top 4 figures) and  FPR (bottom 4 figures), computed on the test set over 1500 runs.  $n=$100, 1000 or 10000; $\alpha=$0.01, 0.05, 0.1, 0.2. The triangle in each boxplot is the mean. For FPR, the red dotted line marks the target FPR of 0.05.}
\label{fig:pac}
\end{figure}

\para{\bf Convergence of relative scoring bias ($\hat{\xi}$) and FPR.}  \rebuttal{Here we present the convergence results in Figure~\ref{fig:pac} for the synthetic dataset in terms of the quantile distribution of $\hat{\xi}$ (computed as the difference of the empirical TPR according to \Eqref{eq:emprelativebias}) between Deep SVDD and Deep SAD and the quantile distribution of Deep SAD's FPR.  Results for other models and three real-world datasets are in Appendix~\ref{section:pac-additional}, and show consistent trends. }


Similar to our theoretical results,  we observe a consistent trend of convergence in FPR and $\hat{\xi}$ as the sample complexity goes up. In particular, as $n$ goes up, FPR converges to the prefixed value of $0.05$ and $\hat{\xi}$ also converges to a certain level.

We also examine the rate of convergence w.r.t to $n$.  Section~\ref{subsec:finite} shows that $n$ required for estimating $\hat{\xi}$ grows in the same order as $\frac{1}{\alpha^2\epsilon^2} \log\frac{1}{\delta}$. That is, the estimation error $\epsilon$ decreases at the rate of $\frac{1}{\sqrt{n}}$; furthermore,
as $\alpha$ increases, $n$ required for estimating $\hat{\xi}$ decreases.  This can be seen from Figure~\ref{fig:pac} (top figure) where at $n=10000$, the variation of $\hat{\xi}$ at $\alpha=0.2$ is 50\% less than that at $\alpha=0.01$.

\vspace{-0.10in}

\section{Impact of Scoring Bias on Anomaly Detection Performance}
\label{sec:eval}
We perform experiments to study the end-to-end impact of relative scoring bias on deep anomaly detection models. Our goal is to understand the type and severity of performance variations caused by different anomaly training sets.

\para{\bf Experiment setup.}
We consider six deep anomaly detection models previously listed in Table~\ref{tab:modeldata}, and three real-world datasets: Fashion-MNIST, Statlog and Cellular Spectrum Misuse.   For each dataset, we build normal data by choosing a single class (e.g.,  \texttt{top} in Fashion-MNIST), and treat other classes as abnormal classes. 
From those abnormal classes, we pick a single class as the abnormal training data, and the rest as the abnormal test data on which we test separately.  We then train $\model_0 := (\score_{\model_0}, \threshold_{\model_0})$, a \rebuttal{semi-supervised} anomaly detector using normal training data,  and $\model_s := (\score_{\model_s}, \threshold_{\model_s})$ a \rebuttal{supervised} anomaly detector using both normal and abnormal training data.  We follow the original paper of each model to implement the training process. 
Detailed descriptions on datasets and training configurations are listed in Appendix~\ref{sec:datasets}.

We evaluate the potential bias introduced by different abnormal training data by comparing the model recall (\TPR) value of both $\model_0$ and $\model_s$ against different abnormal test data.  We define the bias to be upward ($\cuparrow$) if $\TPR(\model_s)>\TPR(\model_0)$, and downward ($\cdownarrow$) if $\TPR(\model_s)<\TPR(\model_0)$.

We group experiments into \rebuttal{three} scenarios:  (1) abnormal training data is visually similar to normal training data; (2) abnormal training data is visually dissimilar to normal training data; \rebuttal{and (3) abnormal training data is a weighted combination of (1) and (2)}.  We compute visual similarity as the $L^2$ distance, which are listed in Appendix~\ref{sec:addeval}.

We observe similar trends across all three datasets and all six anomaly detection models. For brevity, we summarize and illustrate our examples by examples of two models~(Deep SVDD as $\model_0$ and Deep SAD as $\model_s$), and two datasets~(Fashion-MNIST and Cellular Spectrum Misuse).  We report full results on all the models and datasets in Appendix~\ref{sec:addeval}.

\para{\bf Scenario 1: Abnormal training data visually similar to normal training data.} In this scenario, the use of abnormal data in model training does improve detection on the abnormal training class, but also creates considerable performance changes, both upward and downward, for other test classes. The change direction depends heavily on the similarity of the abnormal test data to the training abnormal data.  The model performance on test data similar to the training abnormal data moves {\em upward} significantly while that on test data dissimilar to the training abnormal moves {\em downward} significantly.

\begin{figure}[ht]
\begin{subfigure}[b]{0.20\textwidth}
         \centering
         \includegraphics[width=\textwidth]{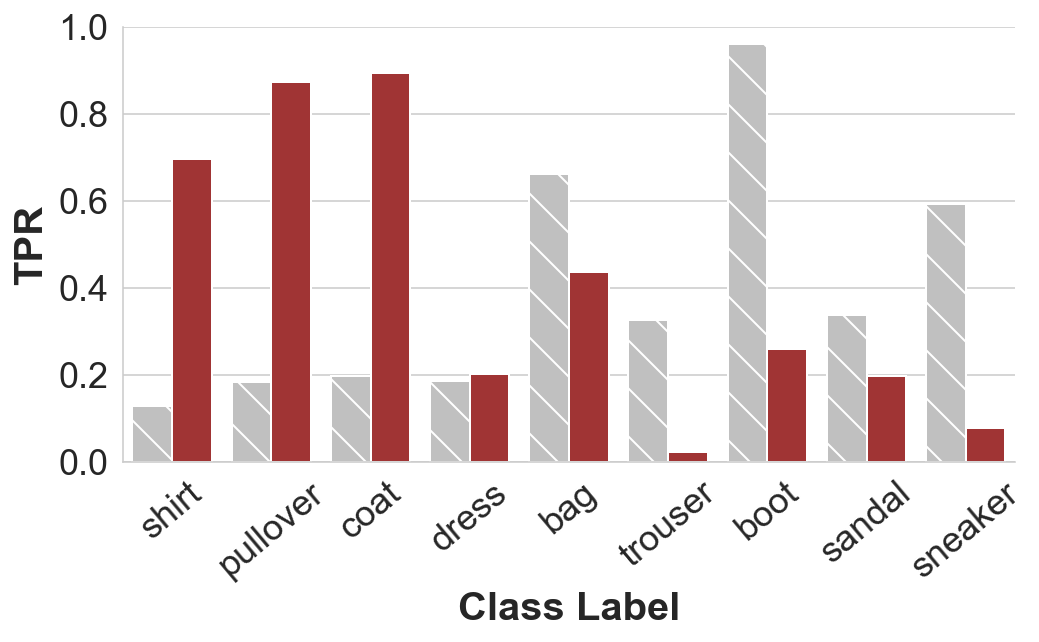}
         \vspace{-0.20in}
         \caption{{\scriptsize Fashion-MNIST: Scenario 1}}
         \end{subfigure}
     \hfill
     \begin{subfigure}[b]{0.20\textwidth}
         \centering
    \includegraphics[width=\textwidth]{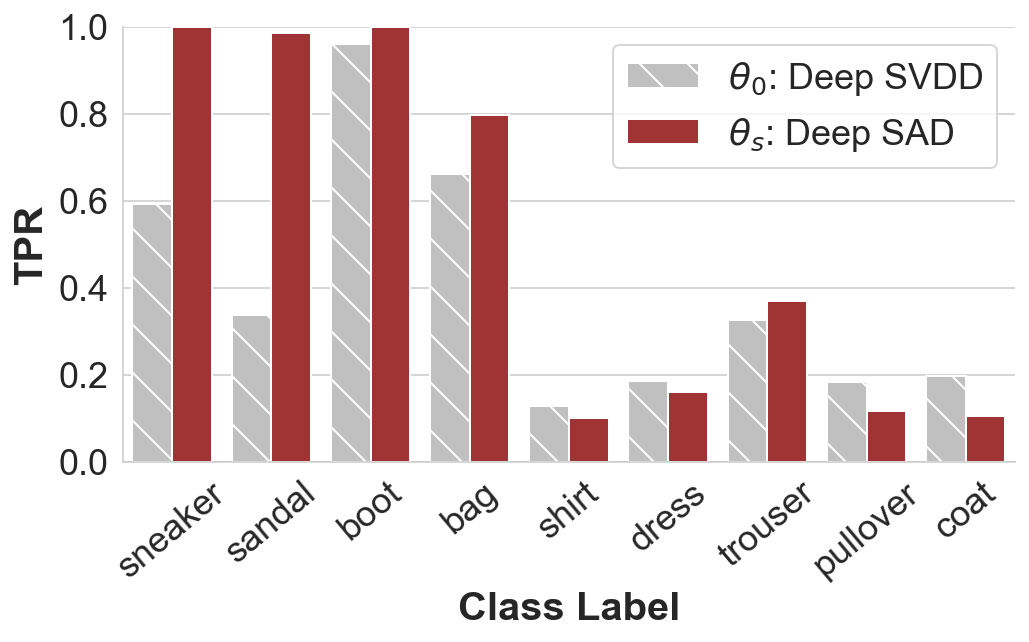}
    \vspace{-0.20in}
          \caption{{\scriptsize Fashion-MNIST: Scenario 2}}
         \end{subfigure}

\begin{subfigure}[b]{0.20\textwidth}
         \centering
         \includegraphics[width=\textwidth]{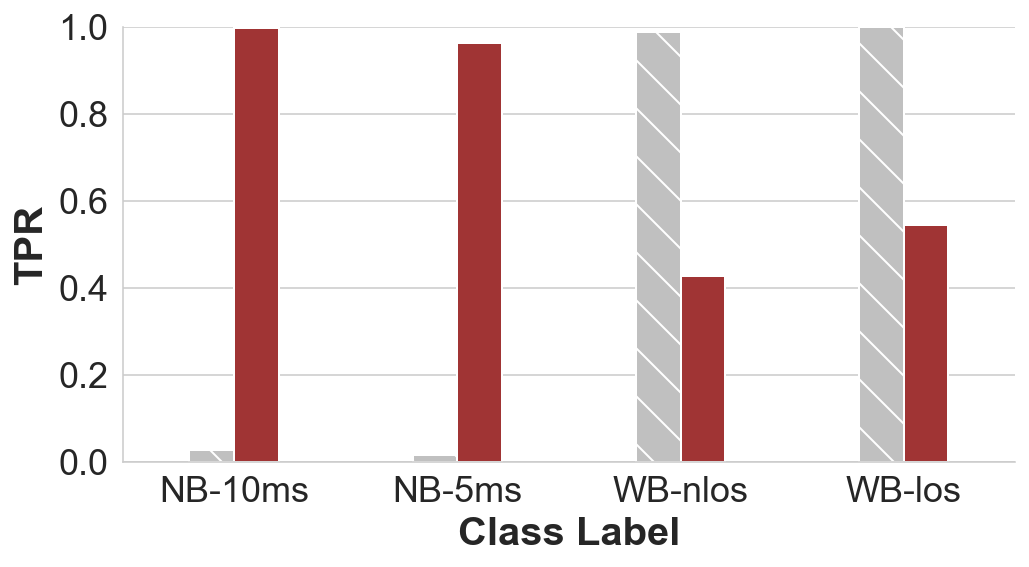}
         \vspace{-0.20in}
         \caption{{\scriptsize Spectrum Misuse: Scenario 1}}
         \end{subfigure}
     \hfill
     \begin{subfigure}[b]{0.20\textwidth}
         \centering
    \includegraphics[width=\textwidth]{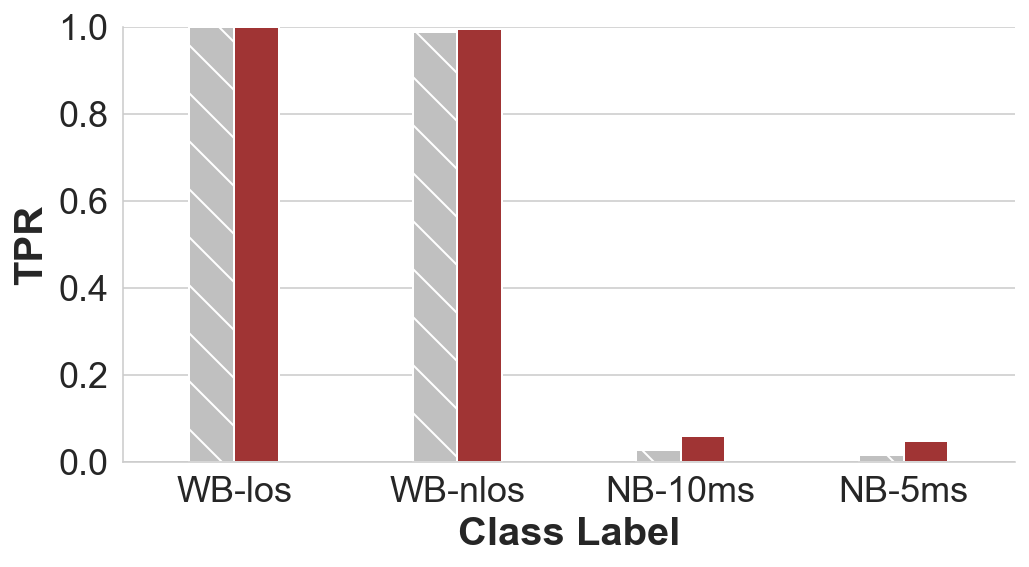}
    \vspace{-0.20in}
          \caption{{\scriptsize Spectrum Misuse: Scenario 2}}
         \end{subfigure}
\vspace*{-2mm}
\caption{\rebuttal{Model \TPR under Scenario 1 and 2, trained on Fashion-MNIST and Cellular Spectrum Misuse. In each figure, we compare the performance of $\model_0$ = Deep SVDD and  $\model_s$ = Deep SAD when tested on abnormal data. We arrange abnormal test data (by their class label) in decreasing similarity with training abnormal data. The leftmost entry in each figure is the class used for abnormal training. For Fashion-MNIST, the normal data is \texttt{top}; for Cellular Spectrum Misuse, the normal data is \texttt{normal}.}}\vspace{-0.1in}
\label{fig:scenario}
\end{figure}

For Fashion-MNIST, the normal and abnormal training classes are {\tt top} and {\tt shirt}, respectively, which are similar to each other. Figure~\ref{fig:scenario}(a) plots the recalls of model  $\model_0$ and $\model_s$ for all abnormal classes, sorted by their similarity to the training abnormal class ({\tt shirt}).  We see that $\TPR(\model_s)$ on classes similar to {\tt shirt} (e.g., {\tt pullover}) is significantly higher than $\TPR(\model_0)$. But for classes dissimilar from {\tt shirt} (e.g., {\tt boot}),  $\TPR(\model_s)$ is either similar or significantly lower.  \rebuttal{For Cellular Spectrum Misuse, the normal and abnormal training classes are {\tt normal} and {\tt NB-10ms}, respectively. The effect of training bias is highly visible in Figure~\ref{fig:scenario}(c), where $\TPR(\model_s)$ on  {\tt NB-10ms} and {\tt NB-5ms} rises from almost 0 to $>$93\% while $\TPR(\model_s)$ on {\tt WB-nlos} and {\tt WB-los} drops by over 50\%. }

\looseness -1 \para{\bf Scenario 2: Abnormal training data visually dissimilar to normal training data.}  Like in Scenario 1,  abnormal training examples improve the detection of abnormal data belonging to the training class and those similar to the training class. Different from Scenario 1, there is little downward changes at abnormal classes dissimilar to the training abnormal.

This is illustrated using another Fashion-MNIST example in Figure~\ref{fig:scenario}(b). While the normal training class is still {\tt top}, we use a new abnormal training class of {\tt sneaker} that is quite dissimilar from {\tt top}. $\TPR(\model_s)$ on {\tt sneaker}, {\tt sandal}, {\tt boot} and {\tt bag} are largely elevated to 0.8 and higher, while $\TPR(\model_s)$ on other classes are relatively stable. \rebuttal{Finally, the same applies to another example of Cellular Spectrum Misuse in Figure~\ref{fig:scenario}(d) where the abnormal training class is {\tt WB-los}, which is quite different from the normal data. In this case, we observe little change to the model recall.}

\rebuttal{\para{\bf Scenario 3: Mixed abnormal training data.} } \rebuttal{We run three configurations of group training on Fashion-MNIST (normal: {\tt top}; abnormal: {\tt shirt} \& {\tt sneaker}) by varying weights of the two abnormal classes in training. The detailed results for each weight configuration are listed in  Appendix~\ref{sec:addconfig}. Overall, the use of group training does improve the model performance, \yzy{but can still introduce downward bias on some classes}. However, under all three weight configurations, we observe a consistent pattern of downward bias for an abnormal test class (\texttt{trouser}) and upward bias for most other abnormal classes. Specifically, \texttt{trouser} is relatively more dissimilar to both training abnormal classes.
}


\para{\bf Summary.} Our empirical study shows that training with biased anomalies can have significant impact on deep anomaly detection, \rebuttal{especially on {\em whether using labeled anomalies in training would help detect unseen anomalies}.  When the labeled anomalies are similar to the normal instances, the trained model will likely face large performance degradation on unseen anomalies {\em different} from the labeled anomalies, but improvement on those {\em similar} to the labeled anomalies. When the labeled anomalies are dissimilar to the normal instances, the supervised model is more useful than its semi-supervised version. Such difference is likely because different types of abnormal data affect the training distribution (thus the scoring function) differently. In particular, when the labeled anomalies are similar to the normal data, they lead to large changes to the scoring function and affect the detection of unseen anomalies ``unevenly''. Our results suggest that model trainers must treat labeled anomalies with care.}

\section{Conclusions and Future Work}
\looseness -1 To the best of our knowledge, our work provides the first formal analysis on how additional labeled anomalies in training affect deep anomaly detection. We define and formulate the impact of training bias on anomaly detector's recall (or \TPR) as the relative {\em scoring bias} of the detector when comparing to a baseline model. We then establish finite sample rates for estimating the relative scoring bias for supervised anomaly detection, and empirically validate our theoretical results on both synthetic and real-world datasets.  We also empirically study how such relative scoring bias translates into variance in detector performance against different types of unseen anomalies, and demonstrate scenarios in which additional labeled anomalies can be useful or harmful. As future work, we will investigate how to construct novel deep anomaly detection models by exploiting upward scoring bias while avoiding downward scoring bias, especially when one can actively collect/ synthesize new labeled anomalies.

\subsubsection*{Acknowledgements}
This work is supported in part by NSF grants CNS-1949650 and CNS-1923778, and a C3.ai DTI Research Award 049755. Any opinions, findings, and conclusions or recommendations expressed in this material are those of the authors and do not necessarily reflect the views of any funding agencies.

\bibliographystyle{named}
\bibliography{anomaly}

\onecolumn
\appendix

\section{Proof of \corref{cor:infinite-gaussian}}
\label{sec:proof-cor2}
\begin{proof}[Proof of \corref{cor:infinite-gaussian}]
Assuming the score functions are Gaussian distributed, we can denoted $F_{0}(s)$ as $\Phi( \frac {s - \mu_{0}} {\sigma_{0}} )$, $\tilde{F}_{0}(s)$ as $\Phi( \frac {s - \tilde{\mu}_{0}} {\tilde{\sigma}_{0}} )$, $F_{a}(s)$ as $\Phi( \frac {s - \mu_{a}} {\sigma_{a}} )$, and $\tilde{F}_{a}(s)$ as $\Phi( \frac {s - \tilde{\mu}_{a}} {\tilde{\sigma}_{a}} )$.

Therefore, we have $\Delta_{0} = |(\sigma_{0}\Phi ^{-1}(q)+\mu_{0})-(\tilde{\sigma}_{0}\Phi ^{-1}(q)+\tilde{\mu}_{0})|$.

Thus,
$$\begin{aligned} \xi _{N} &:=\tilde{r} - r  \\ &= F_{a}(F_{0}^{-1}(q)) - \tilde{F}_{a}(\tilde{F_{0}^{-1}(q)}) \\ &= \Phi(\frac{\sigma_{0}\Phi ^{-1}(q)}{\sigma_{a}} + \frac{\mu_{0} - \mu_{a}}{\sigma_{a}}) - \Phi(\frac{\tilde{\sigma}_{0}\Phi ^{-1}(q)}{\tilde{\sigma}_{a}} + \frac{\tilde{\mu}_{0} - \tilde{\mu}_{a}}{\tilde{\sigma}_{a}}) \end{aligned}$$.
\end{proof}

\section{Proof of \thmref{thm:scomplx}}
\label{sec:proofthm}
\begin{proof}[Proof of \thmref{thm:scomplx}]
Our proof builds upon and extends the analysis framework of \cite{liu2018open}, which relies on one key result from \cite{massart1990},
\begin{align}\label{eq:tail-cdf}
\Pr{\sqrt{n} \sup_x |\hat{F}(x) - F(x)| > \lambda} \leq 2\exp(-2\lambda^2).
\end{align}

Here, $\hat{F}(x)$ is the empirical CDF calculated from $n$ samples.
Given a fixed threshold function, \cite{liu2018open} showed that it required
\begin{align}
    n > \frac{1}{2\epsilon_1^2} \log \frac{2}{1-\sqrt{1-\delta}} \cdot \left(\frac{2-\alpha}{\alpha} \right)^2 \label{eq:liu-n}
\end{align}
examples, in order to guarantee $|\hat{F}_a(x) - F_a(x)| \leq \epsilon_1$ with probability at least $1-\delta$ (recall that $\alpha$ denotes the fraction of abnormal data among the $n$ samples).


\rebuttal{Here we note that our proof relies on a novel adaption of \Eqref{eq:tail-cdf}, which allows us to extend our analysis to the convergence of quantile functions}.

To achieve this goal, we further assume the Lipschitz continuity for the CDFs/quantile functions:
\begin{align}
    |F_a(x)-F_a(x')| &\leq \ell_a |x-x'|  \label{eq:tar-ab}\\
    |F'_a(x)-F'_a(x')| &\leq \ell'_a |x-x'| \label{eq:ref-ab}\\
    |{F_0}^{-1}(x)-{F_0}^{-1}(x')| &\leq \ell_0^- |x-x'| \label{eq:tar-normal}\\
    |{F'_0}^{-1}(x)-{F'_0}^{-1}(x')| &\leq {\ell'_0}^- |x-x'| \label{eq:ref-normal}
\end{align}
where $\ell_a, \ell'_a, \ell_0^-, {\ell'_0}^-$ are the Lipschitz constants for $F_a, F'_a, F_0^{-\ycnote{1}}, {F'_0}^{-\ycnote{1}}$, respectively. Combining the above inequalities \eqref{eq:tar-normal} with \eqref{eq:tail-cdf}, we obtain
\begin{align}
    & \Pr{\sup_q\left\vert \hat{F}_0^{-1}(q) - {F}_0^{-1}(q) \right\vert \geq \frac{\lambda}{\sqrt{n_0}}} \nonumber\\
    \leq &
    \Pr{\sup_q\left\vert F_0\left(\hat{F}_0^{-1}(q)\right) - F_0\left({F}_0^{-1}(q)\right) \right\vert \geq \frac{\lambda}{\sqrt{n_0}\ell_0^-} }.\label{eq:quantile-bound}
\end{align}

Let $q = \hat{F}_0(x)$, then \eqref{eq:quantile-bound} becomes
\begin{align}
    &\Pr{\left\vert F_0\left(\hat{F}_0^{-1}(\hat{F}_0(x))\right) - F_0\left({F}_0^{-1}(\hat{F}_0(x))\right) \right\vert \geq \frac{\lambda}{\sqrt{n_0}\ell_0^-} } \nonumber\\
    =&
    \Pr{\left\vert F_0\left(x\right) - \hat{F}_0(x)) \right\vert \geq \frac{\lambda}{\sqrt{n_0}\ell_0^-} } \nonumber
    \\
    \leq&  2e^{-\frac{2}{n_0}\left(\frac{\lambda}{\ell_0^-}\right)^2}. \nonumber 
\end{align}

Therefore, in order for $\Pr{\sup_q\left\vert \hat{F}_0^{-1}(q) - {F}_0^{-1}(q) \right\vert \geq \epsilon_2} \leq \delta$ to hold, it suffices to set
\begin{align*}
    n_0 = n (1-\alpha) > \frac{1}{2}\frac{1}{\epsilon_2^2} \frac{1}{{\ell^-_0}^2} \log\frac{2}{\delta}.
\end{align*}

Furthermore, combining \eqref{eq:tail-cdf}, \eqref{eq:liu-n}, and \eqref{eq:tar-ab}, we get
\begin{align*}
    \hat{F}_a\left(\tau_{n}\right) \leq F_a\left(\tau\right) + (\tau - \tau_n) \ell_a + \epsilon_1. 
\end{align*}

Subtitute $\tau_n = \hat{F}_0^{-1}(q)$, and $\tau = {F}_0^{-1}(q)$ in the above inequality, and set $\epsilon_1 = \frac{\epsilon}{4}$, $\epsilon_2 = \frac{\epsilon}{4\ell_a}$, we get
\begin{align*}
    \left|\hat{F}_a\left(\hat{F}_0^{-1}(q)\right) - F_a\left({F}_0^{-1}(q)\right)\right| \leq \epsilon/2. 
\end{align*}

Similary, we repeat the same procedure for $s'$, and can get
\begin{align*}
    \left|\hat{F'}_a \left({\hat{F}_0}^{'-1}(q)\right) -  F'_a\left({{F}'_0}^{-1}(q)\right)\right|\leq \epsilon/2. 
\end{align*}
with probability at least $1-\delta$.

Therefore, with $n \geq \frac{8}{\epsilon^2} \cdot \left( \log \frac{2}{1-\sqrt{1-\delta}} \cdot \left(\frac{2-\alpha}{\alpha} \right)^2 + \log \frac{2}{\delta} \cdot \frac{1}{1-\alpha} \left(\left( \frac{\ell_a}{\ell_0^-} \right)^2 + \left( \frac{\ell'_a}{{\ell'_0}^-} \right)^2\right) \right)$ examples we can get $|\hat\xi - \xi| \leq \epsilon$ with probability at least $1-\delta$.

\camera{Note that our results can be easily extended to the setting where the goal
is to minimize FPR subject to a given TPR. The threshold setting, either by fixing TPR or fixing FPR (line 3--4 of \algref{alg:twostage-part2}), is independent of the scoring function (line 1--2 of \algref{alg:twostage-part2}). Thus, it has no impact on the distributions of anomaly scores and our proof technique applies. Specifically, assume the target TPR is $1-q$ and let $\xi\left(s, s^{\prime}\right) := \operatorname{FPR}\left(s^{\prime}, \tau^{\prime}\right)-\operatorname{FPR}(s, \tau)$, we could then modify \propref{prop:infinite} such that  $\xi\left(s, s^{\prime}\right) = F_{0}\left(F_{a}^{-1}(q)\right)-F_{0}^{\prime}\left(F_{a}^{\prime-1}(q)\right)$. Here the only difference from the original setting is that positions of $F_{0}$ and $F_{a}$ interchanges, thus the key steps of the proof remain the same.}

\end{proof}

\section{Three Real-World Datasets and Training Configurations}
\label{sec:datasets}

\para{\bf Fashion-MNIST.} This dataset \citep{xiao2017fashionmnist} is a collection of 70K grayscale images on fashion objects (a training set of 60K examples and a test set of 10K examples), evenly divided into 10 classes (7000 images per class). Each image is of 28 pixels in height and 28 pixels in width, and each pixel-value is  an integer between 0 and 255.  The 10 classes are denoted as \texttt{top}, \texttt{trouser}, \texttt{pullover},  \texttt{dress},  \texttt{coat},  \texttt{sandal},  \texttt{shirt},  \texttt{sneaker},  \texttt{bag},  \texttt{boot}.

To train the anomaly detection models, we pick one class as the normal training class,  another class as the abnormal training class, and the rest as the abnormal testing class. We use the full training set of the normal class (6K), and a random 10\% of the training set of the abnormal training class (600) to train the deep anomaly detection models.  We use the test data of the normal class (1K) to configure the anomaly scoring thresholds to meet a 5\% false positive rate (FPR). We then test the models on the full data of each abnormal testing class, as well as on the untrained fraction of the abnormal training class.

\para{\bf StatLog (Landsat Satellite).} This dataset \citep{Statlogdataset} is a collection of 6,435 NASA satellite images, each of 82 $\times$ 100 pixels, valued between 0 and 255. The six labeled classes are denoted as \texttt{red soil}, \texttt{cotton crop}, \texttt{grey soil}, \texttt{damp grey soil}, \texttt{soil with vegetation stubble}, and \texttt{very damp grey soil}.  Unless specified otherwise, we follow the same procedure to train the models. The normal training data includes 80\% data of the designated class, and the abnormal training data is 10\% of the normal training data in size. Due to the limited amount of the data, we use the full data of the normal data to configure the anomaly scoring thresholds to meet a 5\% FPR. We then test the models on the full data of each abnormal testing class, as well as the untrained fraction of the abnormal training class.

\rebuttal{
\para{\bf Cellular Spectrum Misuse.} This real-world anomaly dataset measures cellular spectrum usage under both normal scenarios and in the presence of misuses (or attacks)~\citep{limobihoc2019}. We obtained the dataset from the authors of~\cite{limobihoc2019}. The dataset includes a large set (100K instances) of real cellular spectrum measurements in the form of {\em spectrogram} (or time-frequency pattern of the received signal). Each spectrogram (instance) is a 125$\times$128 matrix, representing the signal measured over 125 time steps and 128 frequency subcarriers. The dataset includes five classes: {\tt normal} (normal usage in the absence of misuse) and four misuse classes: \texttt{WB-los} (wideband attack w/o blockage), \texttt{WB-nlos} (wideband attack w/ blockage), \texttt{NB-10ms} (narrowband attack) and  \texttt{NB-5ms} (narrowband attack with a different signal). The sample size is 60K for {\tt normal} and 10K for each abnormal class. To train the models, we randomly sample 20K instances from {\tt normal}, 2K instances from one abnormal class, and configure the anomaly score thresholds to meet a 5\% FPR.
}

\yzy{For network structures, we use standard LeNet convultional netwroks for Fashion-MNIST, 3-layer Multilayer Perceptron for StatLog, and 3-layer stacked LSTM for Cellular Spectrum Misuse. The detailed settings for hyperparameters and computing infrastructure follow the original implementation \citep{ruff2020rethinking, yamanaka2019autoencoding, limobihoc2019}.}


\section{Additional Experiment Results of Section~\ref{sec:analysis-casestudy}}
\label{section:pac-additional}
\begin{figure*}[h!]
\centering
 \includegraphics[width=0.60\textwidth]{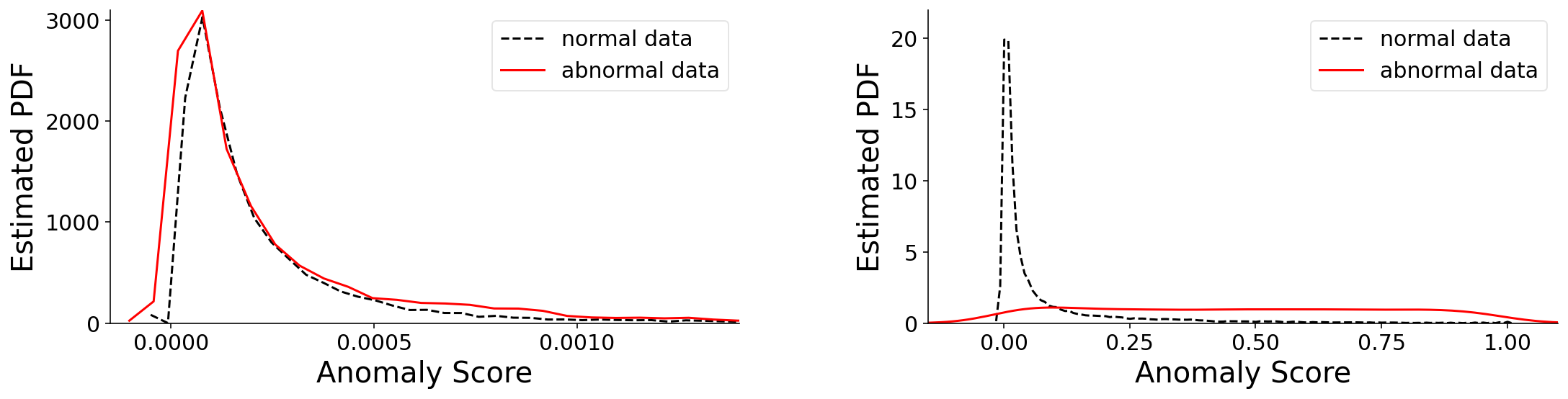}
\caption{Anomaly score distributions for Deep SVDD (left) and HSC (right), trained on the synthetic dataset.}
\label{fig:pdf-hsc}
\end{figure*}

\begin{figure*}[h!]
\centering
 \includegraphics[width=0.60\textwidth]{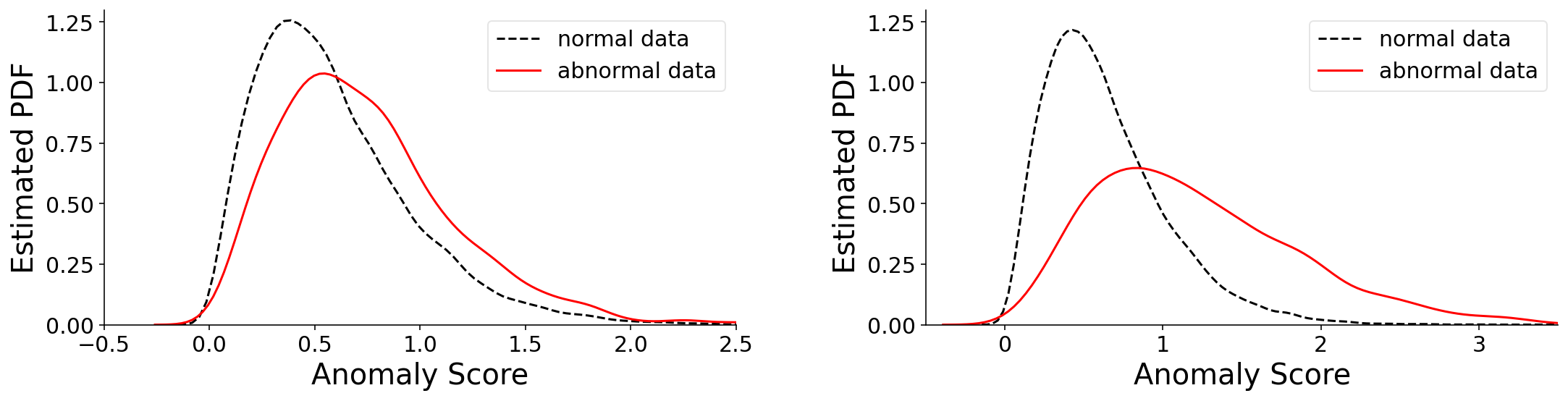}
\caption{Anomaly score distributions for AE (left) and SAE (right), trained on the synthetic dataset.}
\label{fig:pdf-sae}
\end{figure*}

\begin{figure*}[h!]
\centering
 \includegraphics[width=0.60\textwidth]{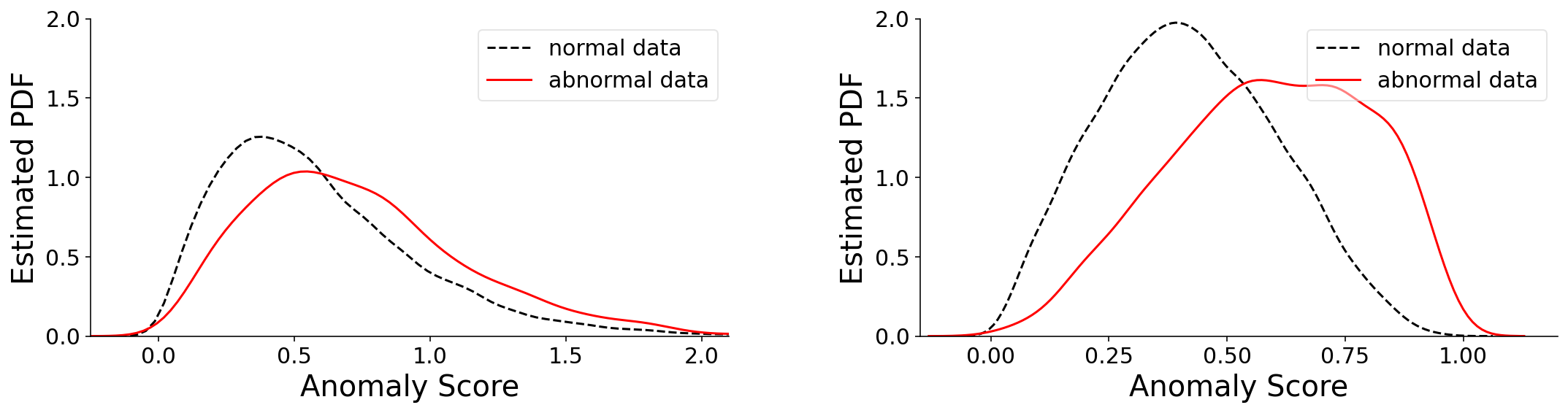}
\caption{Anomaly score distributions for AE (left) and ABC (right),trained on the synthetic dataset.}
\label{fig:pdf-abc}
\end{figure*}

\para{\bf Anomaly score distributions for models trained on the synthetic dataset.} Figure~\ref{fig:pdf-hsc}--\ref{fig:pdf-abc} plot the anomaly score distributions for semi-supervised (left figure) and supervised (right figure) models, trained on the synthetic dataset ($\alpha=0.1$, $n=10000$), estimated by Kernel Density Estimation.

\begin{figure*}[h!]
\centering
 \includegraphics[width=0.60\textwidth]{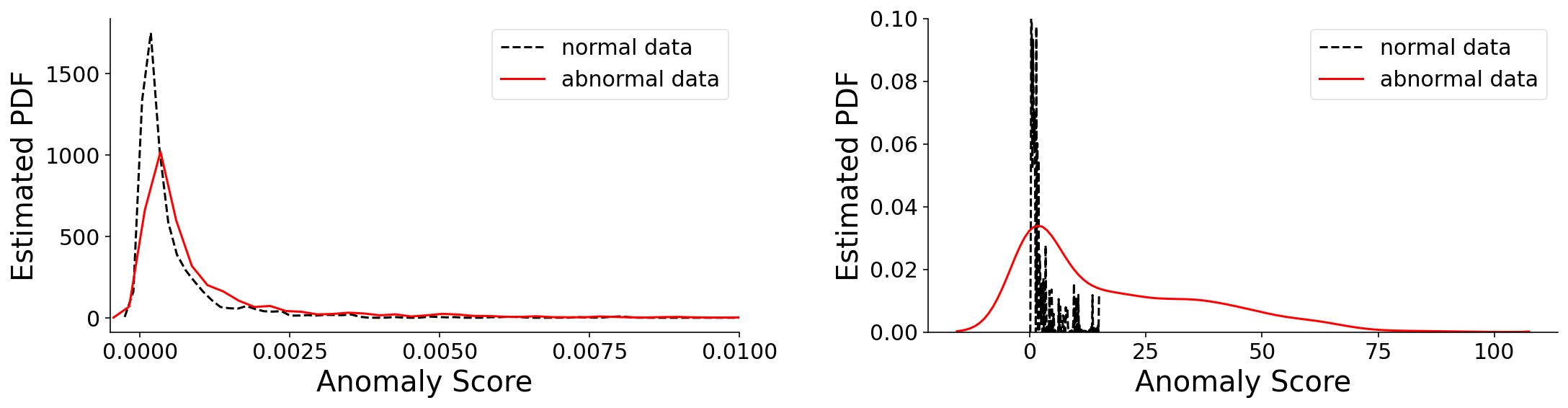}
\caption{Anomaly score distributions for Deep SVDD (left) and Deep SAD (right), trained on Fashion-MNIST (using \texttt{top} as the normal class and \texttt{shirt} as the abnormal class).}
\label{fig:pdf-fmnist}
\end{figure*}

\begin{figure*}[h!]
\centering
\includegraphics[width=0.60\textwidth]{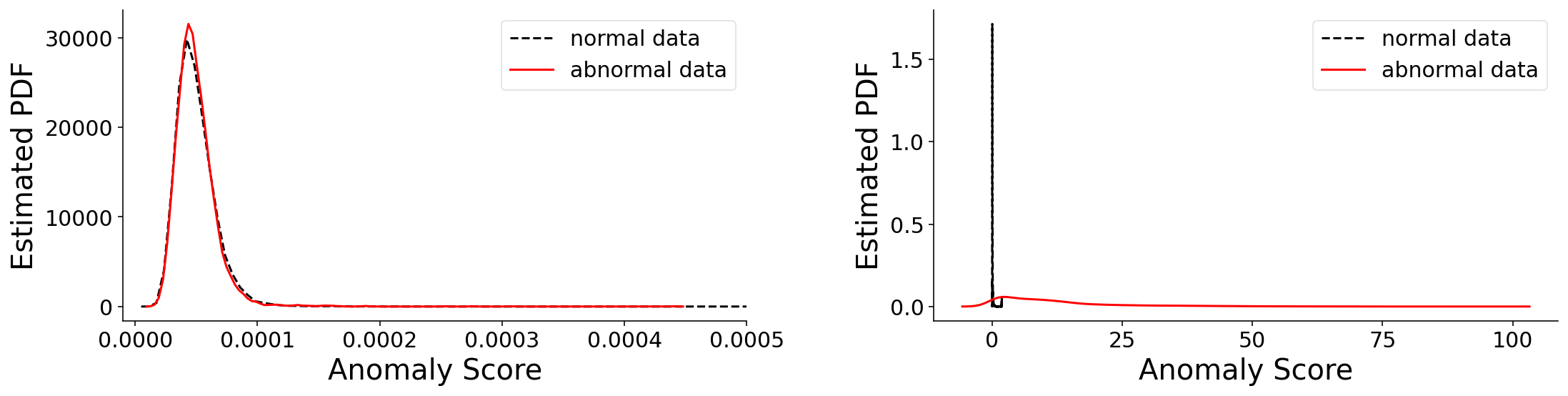}
\caption{Anomaly score distributions for Deep SVDD (left) and Deep SAD (right), trained on Cellular Spectrum Misuse (using \texttt{normal} as the normal class and \texttt{NB-10ms} as the abnormal class).
}
\label{fig:pdf-spectrum}
\end{figure*}

\para{\bf Anomaly score distributions for models trained on real-world datasets.} Figure~\ref{fig:pdf-fmnist} and~\ref{fig:pdf-spectrum} plot the anomaly score distributions for semi-supervised and supervised models, when trained on  Fashion-MNIST and Cellular Spectrum Misuse, respectively.

\newpage

\begin{figure*}[h!]
       \centering
\includegraphics[width=0.8\textwidth]{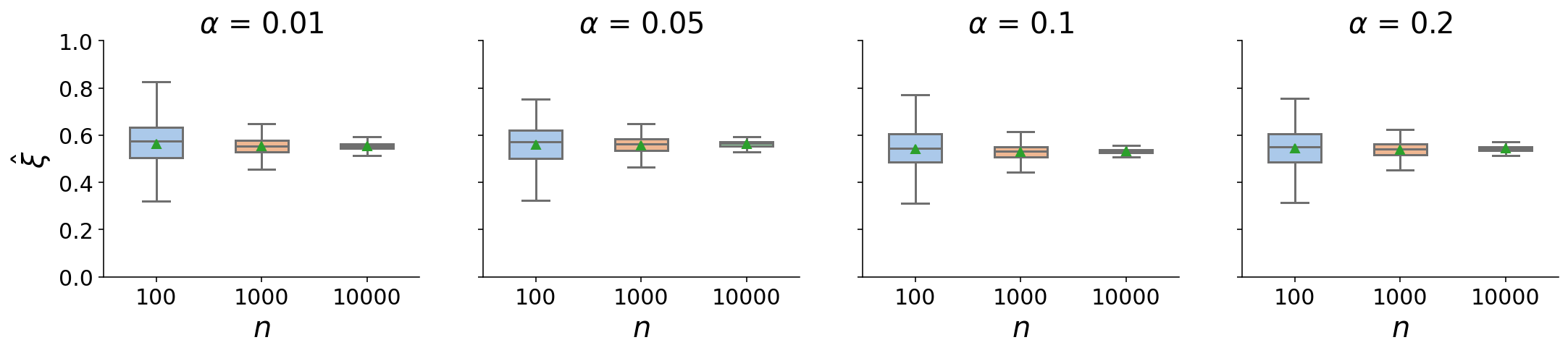}
\includegraphics[width=0.8\textwidth]{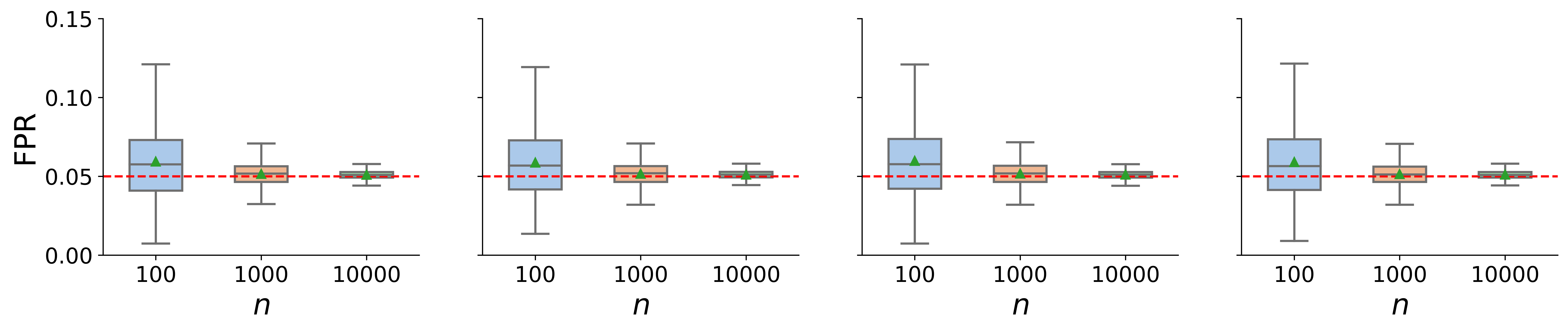}
\caption{The quantile distribution on the synthetic dataset of (top) relative scoring bias $\hat{\xi}$ and (bottom) FPR, computed on the test set over 1500 runs, for Deep SVDD and HSC.}
\label{fig:pac-svdd-hsc}
\end{figure*}

\begin{figure*}[h!]
         \centering
\includegraphics[width=0.8\textwidth]{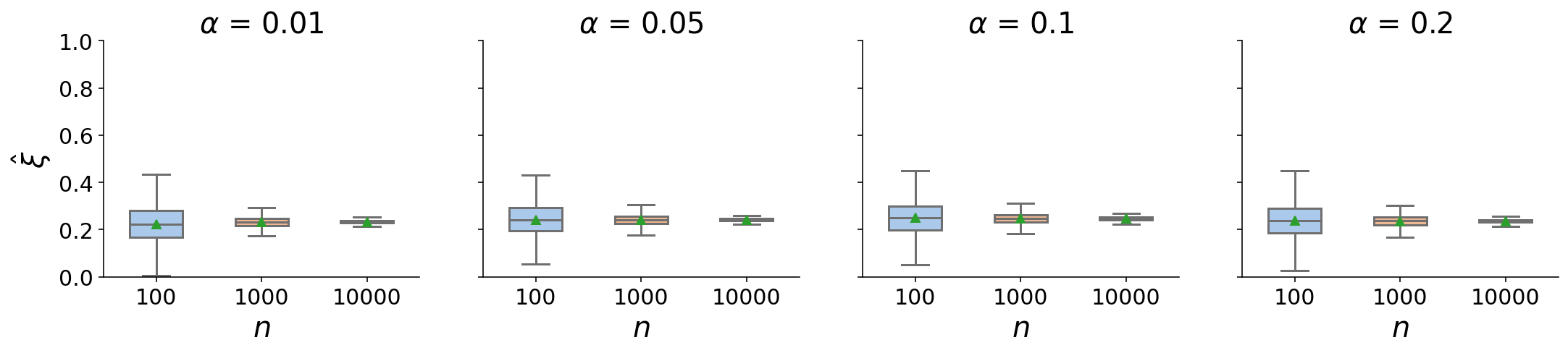}
\includegraphics[width=0.8\textwidth]{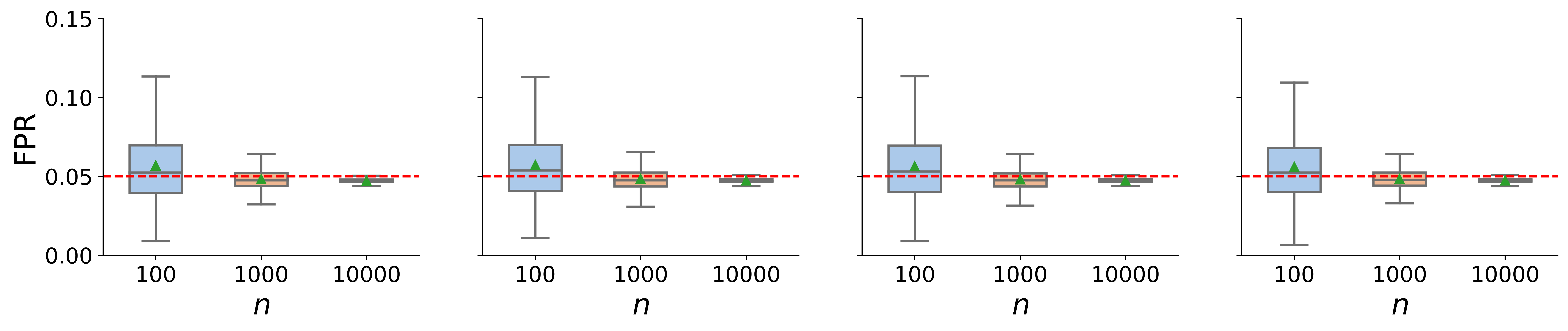}
\caption{The quantile distribution on the synthetic dataset of (top) relative scoring bias $\hat{\xi}$ and (bottom) FPR, computed on the test set over 1500 runs, for AE and SAE.}
\label{fig:pac-ae-sae}
\end{figure*}

\begin{figure*}[h!]
         \centering
\includegraphics[width=0.8\textwidth]{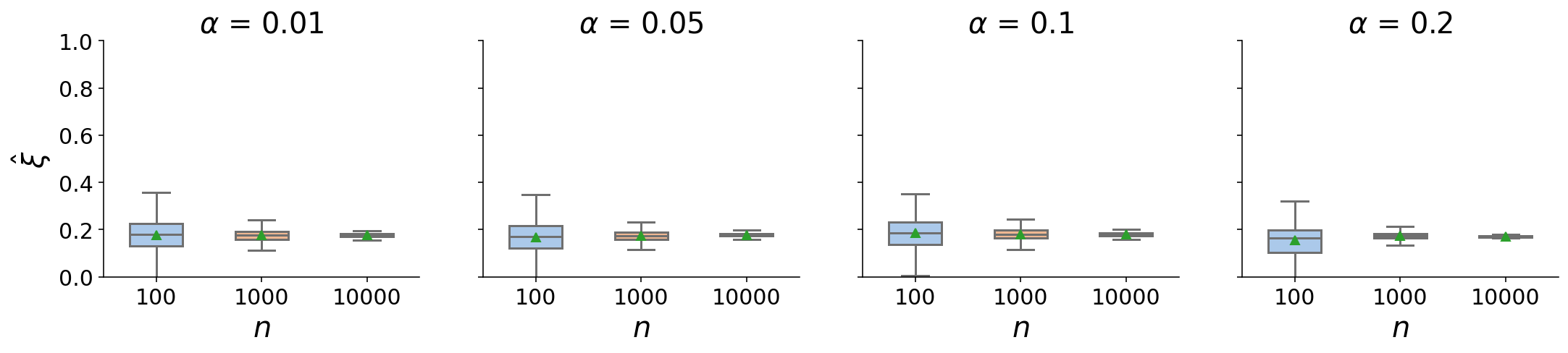}
\includegraphics[width=0.8\textwidth]{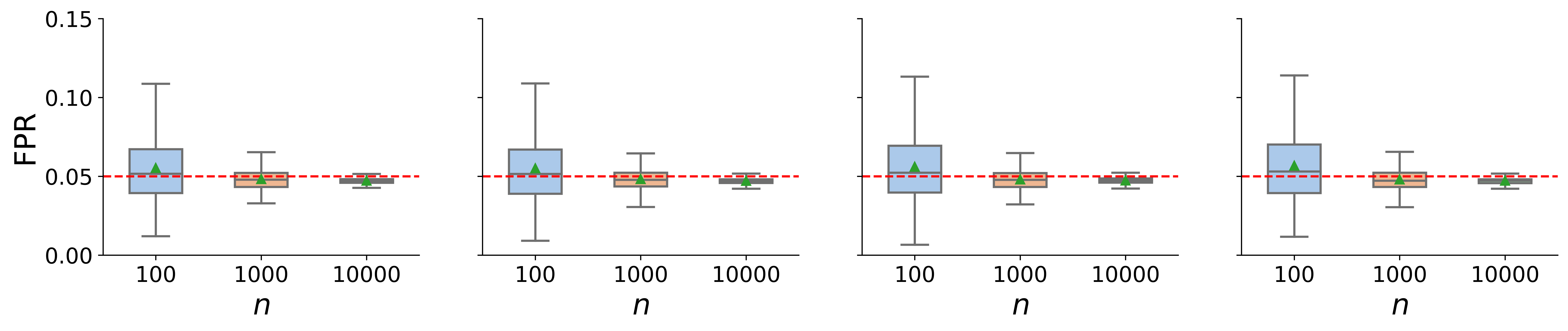}
\caption{The quantile distribution on the synthetic dataset of (top) relative scoring bias $\hat{\xi}$ and (bottom) FPR, computed on the test set over 1500 runs, for AE and ABC.}
\label{fig:pac-ae-abc}
\end{figure*}

\para{\bf Convergence of relative scoring bias $\hat{\xi}$ and FRP on the synthetic dataset.}  We plot in Figure~\ref{fig:pac-svdd-hsc}--\ref{fig:pac-ae-abc} the additional results on (Deep SVDD vs. HSC), (AE vs. SAE), and (AE vs. ABC). Experiment settings are described in Section~\ref{sec:analysis-casestudy}. Overall, they show a consistent trend on convergence.

\newpage

\begin{figure*}[h!]
    \centering
\includegraphics[width=0.8\textwidth]{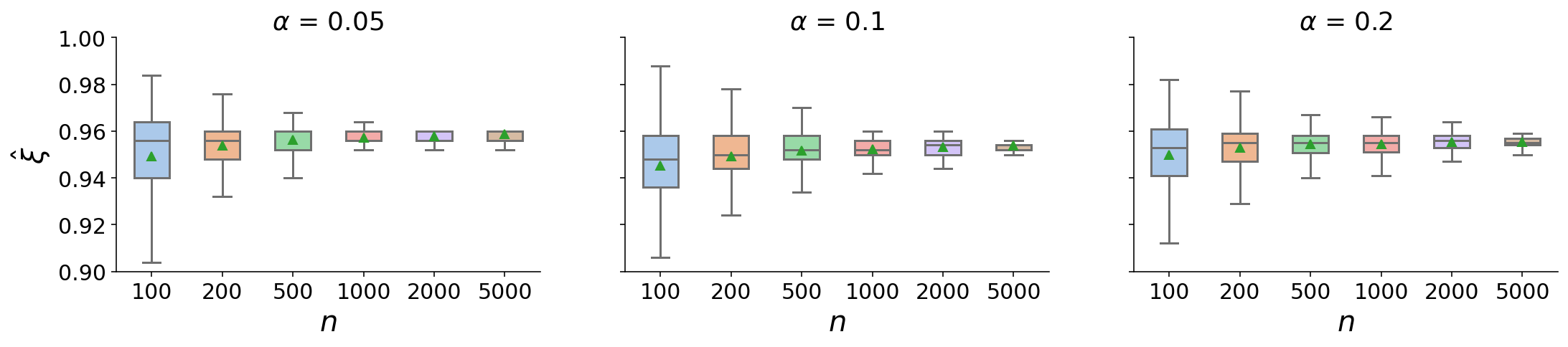}
\includegraphics[width=0.8\textwidth]{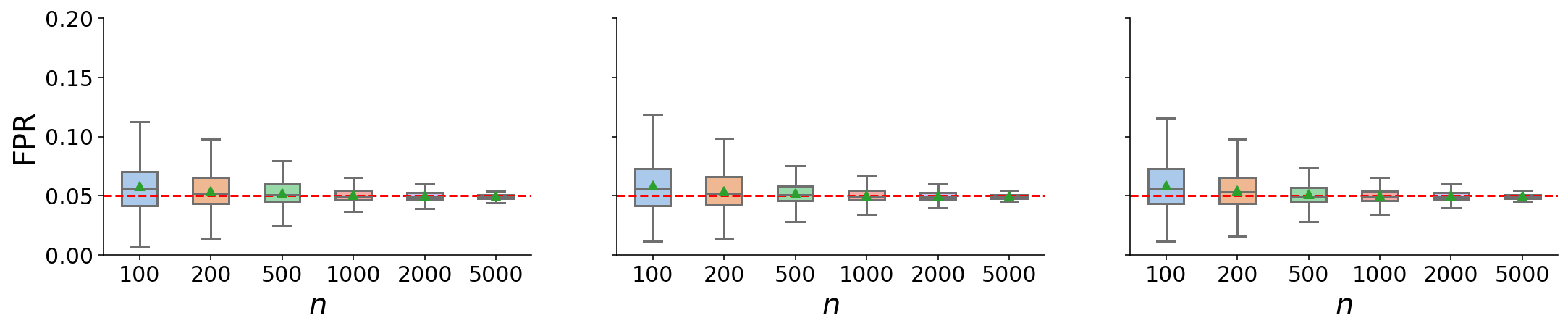}
\vspace*{-1mm}
\caption{The quantile distribution of  relative scoring bias $\hat{\xi}$ (top) and  FPR (bottom), computed on the test set over 1000 runs, for Deep SVDD and Deep SAD trained on Cellular Spectrum Misuse.}
\label{fig:pac-spectrum}
\vspace{-0.1in}
\end{figure*}

\begin{figure*}[h!]
       \centering
\includegraphics[width=0.8\textwidth]{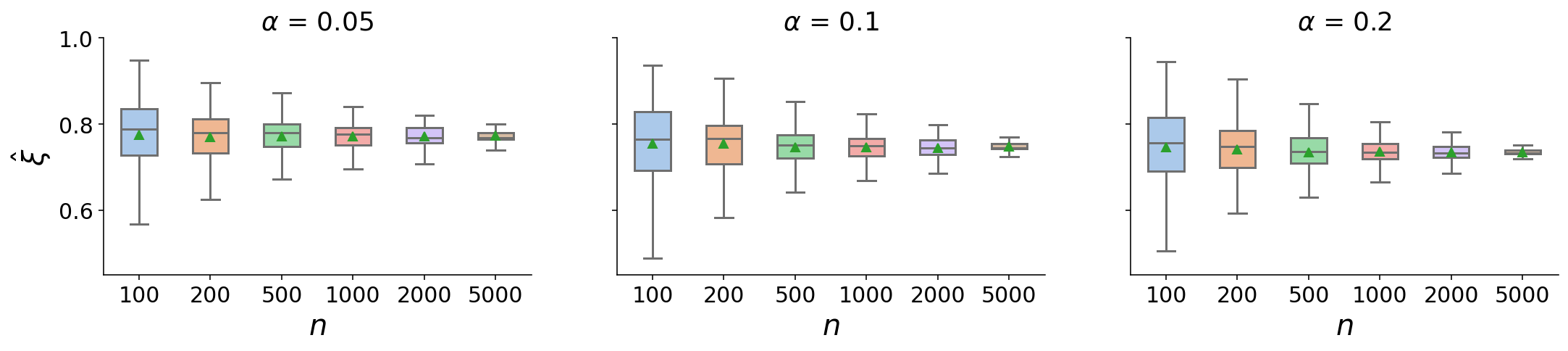}
\includegraphics[width=0.8\textwidth]{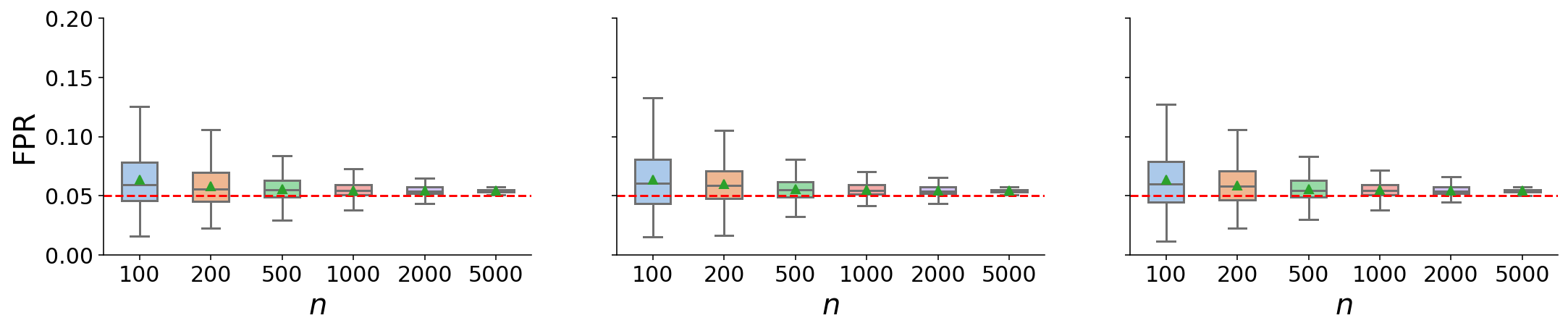}
\caption{The quantile distribution of (top) relative scoring bias $\hat{\xi}$ and (bottom) FPR, computed on the test set over 1000 runs, for Deep SVDD and HSC trained on Cellular Spectrum Misuse.}
\label{fig:sp-pac-svdd-hsc}
\end{figure*}

\begin{figure*}[h!]
         \centering
\includegraphics[width=0.8\textwidth]{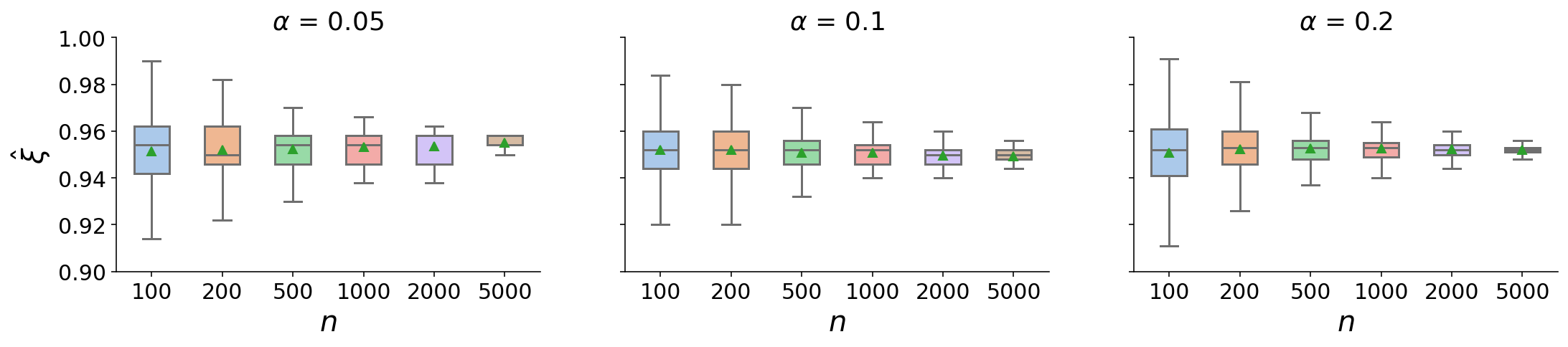}
\includegraphics[width=0.8\textwidth]{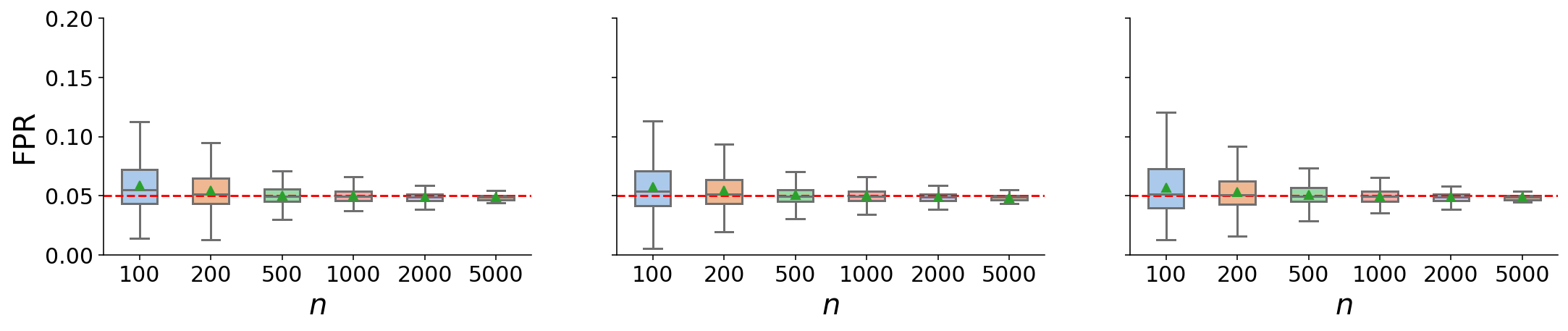}
\caption{The quantile distribution of (top) relative scoring bias $\hat{\xi}$ and (bottom) FPR, computed on the test set over 1000 runs, for AE and SAE trained on Cellular Spectrum Misuse.}
\label{fig:sp-pac-ae-sae}
\end{figure*}

\begin{figure*}[h!]
         \centering
\includegraphics[width=0.8\textwidth]{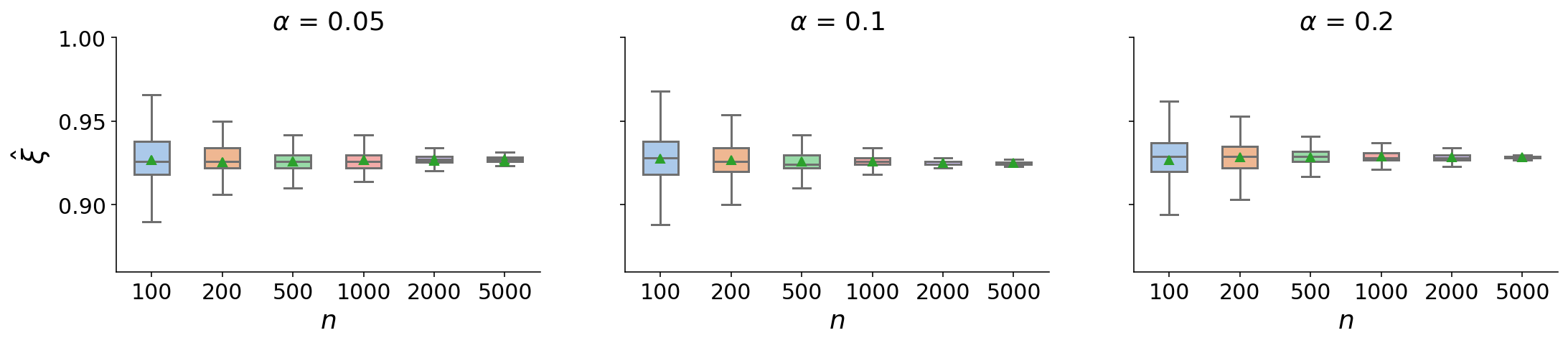}
\includegraphics[width=0.8\textwidth]{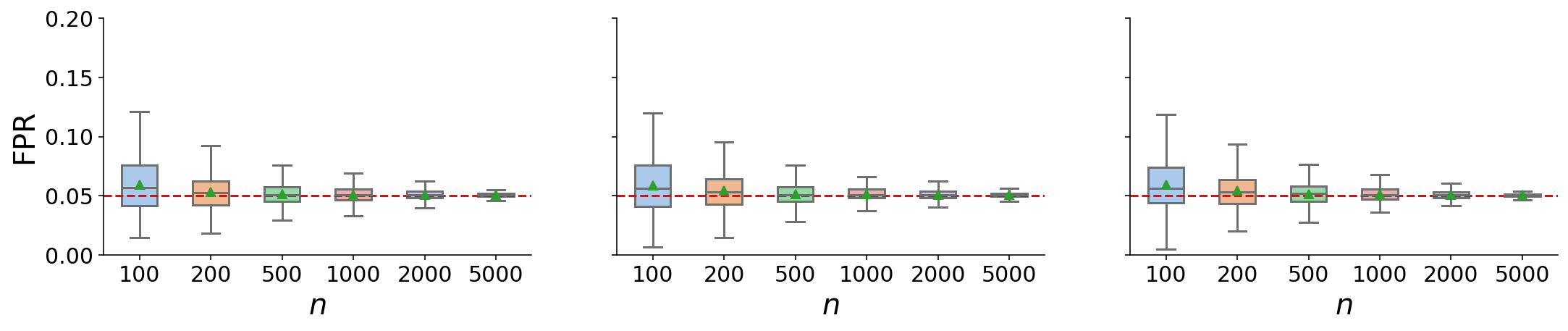}
\caption{The quantile distribution of (top) relative scoring bias $\hat{\xi}$ and (bottom) FPR, computed on the test set over 1000 runs, for AE and ABC trained on Cellular Spectrum Misuse.}
\label{fig:sp-pac-ae-abc}
\end{figure*}

\para{\bf Convergence of relative scoring bias $\hat{\xi}$ and FRP on Cellular Spectrum Misuse.}  We plot in Figure~\ref{fig:pac-spectrum}--\ref{fig:sp-pac-ae-abc} the additional results of (Deep SVDD vs. Deep SAD), (Deep SVDD vs. HSC), (AE vs. SAE), and (AE vs. ABC), when trained on the Cellular Spectrum Misuse dataset. Here we set the normal class as \texttt{normal} and the abnormal class as \texttt{NB-10ms}, and configure the sample size for the training set as 16K and for the test set as 6K, and vary the sample size for the validation set $n$ from {100, 200, 500, 1K, 2K, 5K}. Overall, the plots show a consistent trend on convergence.

\newpage
\newpage

\begin{figure*}[h!]
       \centering
\includegraphics[width=0.8\textwidth]{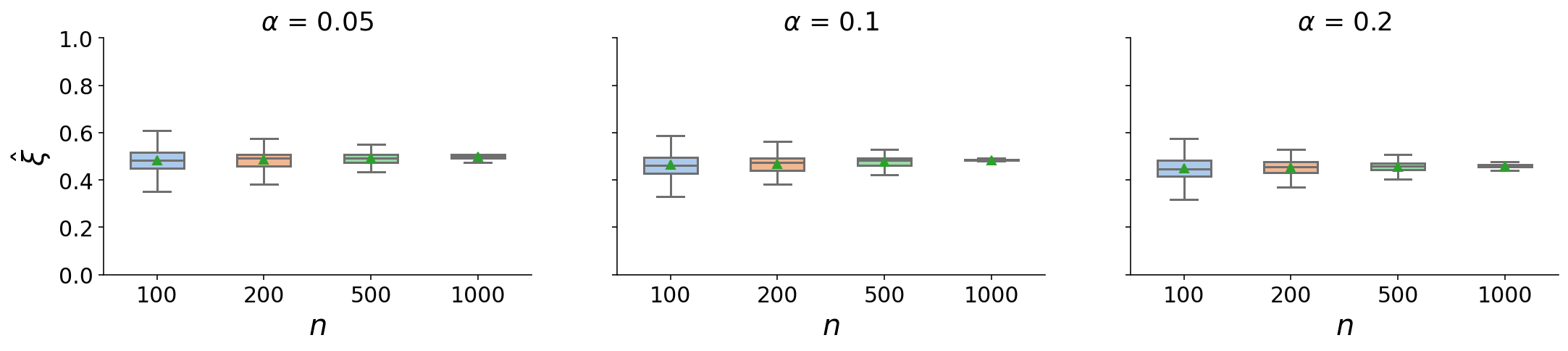}
\includegraphics[width=0.8\textwidth]{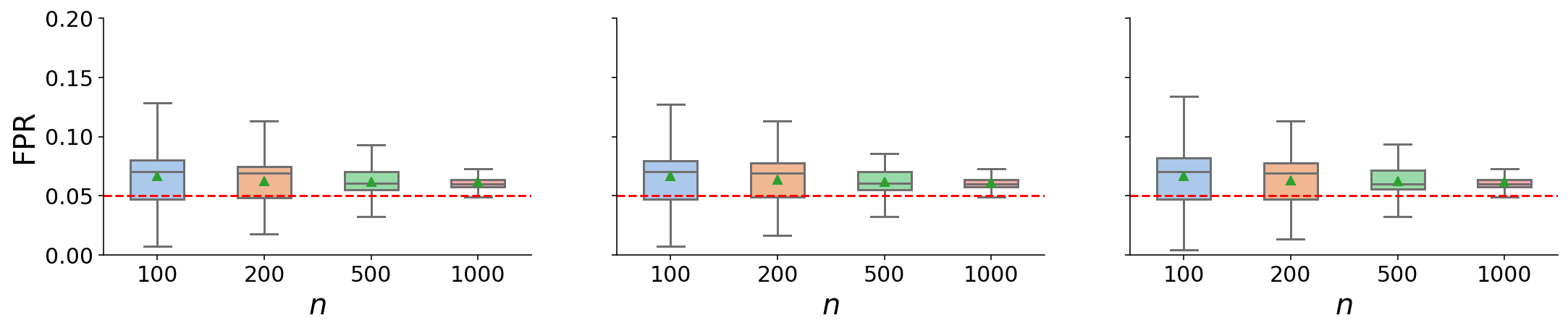}
\caption{The quantile distribution of (top) relative scoring bias $\hat{\xi}$ and (bottom) FPR, computed on the test set over 100 runs, for Deep SVDD and SAD trained on Fashion-MNIST.}
\label{fig:fmnist-pac-svdd-sad}
\end{figure*}

\para{\bf Convergence of relative scoring bias $\hat{\xi}$ and FRP on Fashion-MNIST.}  Figure~\ref{fig:fmnist-pac-svdd-sad} plots the convergence of $\hat{\xi}$ and FRP on Deep SVDD vs. Deep SAD models trained on Fashion-MNIST. The results for other model combinations are consistent and thus omitted. Here we set the normal class as \texttt{top} and the abnormal class as \texttt{shirt}, and configure the sample size for the training set as 3K and for the test set as 2K, and vary the sample size for the validation set $n$ from {100, 200, 500, 1K}. Overall, the plots show a consistent trend on convergence.

\newpage

\begin{figure*}[h]
       \centering
\includegraphics[width=0.8\textwidth]{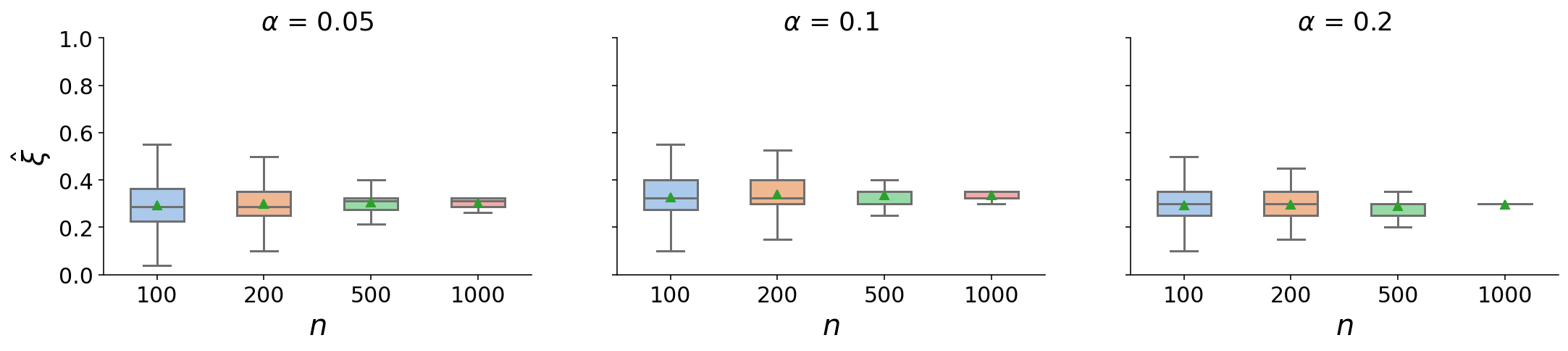}
\includegraphics[width=0.8\textwidth]{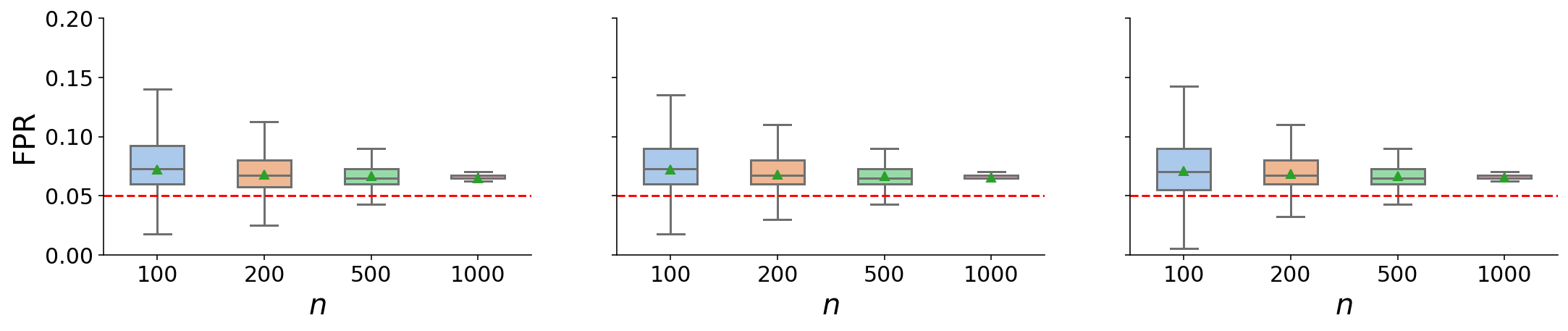}
\caption{The quantile distribution of (top) relative scoring bias $\hat{\xi}$ and (bottom) FPR, computed on the test set over 100 runs, for Deep SVDD and SAD trained on StatLog.}
\label{fig:sati-pac-svdd-sad}
\end{figure*}

\para{\bf Convergence of relative scoring bias $\hat{\xi}$ and FRP on StatLog.}   Figure~\ref{fig:sati-pac-svdd-sad} plots $\hat{\xi}$ and FRP of Deep SVDD vs. Deep SAD trained on the StatLog dataset. The results for other model combinations are consistent and thus omitted.  To maintain a reasonable sample size, we set the normal class to be a combination of \texttt{grey soil}, \texttt{damp grey soil} and \texttt{very damp grey soil}, and the abnormal class to be a combination of \texttt{red soil} and \texttt{cotton crop}. The sample size for the training is 1.2K and  the test set is 1K. We vary the sample size for the validation set $n$ from {100, 200, 500, 1K}. Overall, the plots show a consistent trend on convergence.

\section{Additional Results of Section~\ref{sec:eval}}
\label{sec:addeval}

\para{\bf Scenario 1.} Here the training normal set is very similar to the training abnormal set. We include the detailed recall result (mean/std over 100 runs) of all six models, and three real-life datasets in Table~\ref{tab:fmnist-recall-case1} --~\ref{tab:imagenet-recall-case1}. Across the six models, the two semi-supervised models (trained on normal data) are Deep SVDD and AE; and the rest are supervised models trained on both normal data and the specified abnormal training data. In each table, we report the model recall on all abnormal test classes. These classes are sorted by decreasing similarity to the abnormal training class (measured by $L^2$, small value = visually similar).   Also, $\cuparrow$ indicates that the supervised model has a higher recall than the semi-supervised model;  $\cdownarrow$ indicates the other direction.  Overall we observe both upward and downward bias across the test abnormal classes, and the direction depends on the test abnormal class' similarity to the train abnormal class.

We also observe that when using the reconstruction based models (AE, SAE, ABC), the performance for StatLog is much worse than the hypersphere based models.
\ycnote{This result is in fact consistent with what has been reported in the literature---\citet{ishii2020l0} reports a similar low performance of reconstruction based model on StatLog, which was trained on  normal data (cf Table 4 of \citet{ishii2020l0})}.
We consider this as a result potentially arising from specific latent spaces of data on the reconstruction based models, and leave improvement of these reconstruction models to future work.


\ycnote{It is worth highlighting that although these reconstruction based models demonstrate inferior performance on StatLog when training only on normal data, adding training abnormal set under Scenario 1 does demonstrate a similar behavior to that of the hypersphere based models. When training the SAE model with (biased) anomaly data, we handle the exploding loss issue in a similar way as \citet{ruff2020deep}: For the anomaly class, we consider a loss function which takes the form of $\frac{1}{\text{reconstruction error}}$. We found that this design of loss function easily converges in practice with few loss explosion issues on reconstruction based models.}




\begin{table*}[ht]
\centering
{\small training normal = \texttt{top}, training abnormal = \texttt{shirt}} \\
\resizebox{0.95\textwidth}{!}{%
\begin{tabular}{@{}cccc|ccc |c@{}}
\toprule
\multicolumn{1}{c|}{{Test data}} & \textbf{Deep SVDD} & \textbf{Deep SAD} & \textbf{HSC}  & \textbf{AE} & \textbf{SAE} & \textbf{ABC} & $L^2$ to \texttt{shirt}\\ \hline
\multicolumn{1}{c|}{{\tt shirt}}    & 0.09 $\pm$ 0.01     & 0.71 $\pm$ 0.01 $\cuparrow$    & \multicolumn{1}{c|}{0.70 $\pm$ 0.01 $\cuparrow$}     & 0.12 $\pm$ 0.01        & 0.72 $\pm$ 0.01 $\cuparrow$         & 0.72 $\pm$ 0.01 $\cuparrow$      &0         \\ \hline
\multicolumn{1}{c|}{\tt pullover}   & 0.13 $\pm$ 0.02     & 0.90 $\pm$ 0.01 $\cuparrow$    & \multicolumn{1}{c|}{0.89 $\pm$ 0.01 $\cuparrow$}     & 0.19 $\pm$ 0.02        & 0.84 $\pm$ 0.02 $\cuparrow$         & 0.85 $\pm$ 0.01 $\cuparrow$        &0.01       \\
\multicolumn{1}{c|}{\tt coat}       & 0.14 $\pm$ 0.03     & 0.92 $\pm$ 0.02 $\cuparrow$    & \multicolumn{1}{c|}{0.92 $\pm$ 0.01 $\cuparrow$}      & 0.15 $\pm$ 0.02        & 0.92 $\pm$ 0.02 $\cuparrow$          & 0.92 $\pm$ 0.01 $\cuparrow$            &0.01    \\
\multicolumn{1}{c|}{\tt dress}      & 0.17 $\pm$ 0.03     & 0.24 $\pm$ 0.03 $\cuparrow$    & \multicolumn{1}{c|}{0.24 $\pm$ 0.03 $\cuparrow$}     & 0.11 $\pm$ 0.01        & 0.20 $\pm$ 0.03 $\cuparrow$         & 0.21 $\pm$ 0.03 $\cuparrow$            &0.04   \\
\multicolumn{1}{c|}{\tt bag}        & 0.49 $\pm$ 0.07     & 0.38 $\pm$ 0.08 $\cdownarrow$  & \multicolumn{1}{c|}{0.36 $\pm$ 0.07 $\cdownarrow$}   & 0.70 $\pm$ 0.03        & 0.52 $\pm$ 0.09 $\cdownarrow$       & 0.53 $\pm$ 0.07 $\cdownarrow$          &0.04  \\
\multicolumn{1}{c|}{\tt trouser}    & 0.32 $\pm$ 0.10     & 0.07 $\pm$ 0.04 $\cdownarrow$  & \multicolumn{1}{c|}{0.06 $\pm$ 0.03 $\cdownarrow$}   & 0.59 $\pm$ 0.04        & 0.07 $\pm$ 0.04 $\cdownarrow$       & 0.16 $\pm$ 0.07 $\cdownarrow$          &0.06\\
\multicolumn{1}{c|}{\tt boot}       & 0.92 $\pm$ 0.03     & 0.29 $\pm$ 0.15 $\cdownarrow$  & \multicolumn{1}{c|}{0.27 $\pm$ 0.16 $\cdownarrow$}   & 0.98 $\pm$ 0.02        & 0.90 $\pm$ 0.09 $\cdownarrow$       & 0.90 $\pm$ 0.08 $\cdownarrow$          &0.08   \\
\multicolumn{1}{c|}{\tt sandal}     & 0.30 $\pm$ 0.04     & 0.26 $\pm$ 0.08 $\cdownarrow$    & \multicolumn{1}{c|}{0.26 $\pm$ 0.12 $\cdownarrow$}   & 0.82 $\pm$ 0.02        & 0.46 $\pm$ 0.10 $\cdownarrow$       & 0.56 $\pm$ 0.09 $\cdownarrow$          &0.09     \\
\multicolumn{1}{c|}{\tt sneaker}    & 0.55 $\pm$ 0.09     & 0.12 $\pm$ 0.10 $\cdownarrow$  & \multicolumn{1}{c|}{0.14 $\pm$ 0.12 $\cdownarrow$}   & 0.74 $\pm$ 0.09        & 0.47 $\pm$ 0.19 $\cdownarrow$       & 0.46 $\pm$ 0.18 $\cdownarrow$          &0.10    \\ \bottomrule
\end{tabular}%
}
\caption{The model TPR under scenario 1, Fashion-MNIST. The normal class \texttt{top} is similar to the abnormal training class \texttt{shirt}. Their $L^2$ distance = 0.02.}
\label{tab:fmnist-recall-case1}
\end{table*}

\begin{table*}[ht]
\vspace{-0.1in}
\centering
{\small training normal = \texttt{very damp grey soil}, training abnormal = \texttt{damp grey soil}} \\
\resizebox{0.99\textwidth}{!}{%
\begin{tabular}{@{}cccc|ccc |c@{}}
\toprule
\multicolumn{1}{c|}{{Test data}} & \textbf{Deep SVDD} & \textbf{Deep SAD} & \textbf{HSC}  & \textbf{AE} & \textbf{SAE} & \textbf{ABC} & $L^2$ to \texttt{damp grey soil}\\ \hline
\multicolumn{1}{c|}{{\tt damp grey soil}}    & 0.12 $\pm$ 0.05          & 0.81 $\pm$ 0.02 $\cuparrow$    & \multicolumn{1}{c|}{0.80 $\pm$ 0.02 $\cuparrow$}     & 0.00 $\pm$ 0.00        & 0.08 $\pm$ 0.02 $\cuparrow$        & 0.01 $\pm$ 0.01 $\cdownarrow$       & 0         \\ \hline
\multicolumn{1}{c|}{\tt red soil}            & 0.67 $\pm$ 0.16          & 0.92 $\pm$ 0.05 $\cuparrow$    & \multicolumn{1}{c|}{0.91 $\pm$ 0.05 $\cuparrow$}     & 0.00 $\pm$ 0.00        & 0.00 $\pm$ 0.00 $\keep$                   & 0.00 $\pm$ 0.00 $\keep$                 & 4.39       \\
\multicolumn{1}{c|}{\tt grey soil}           & 0.45 $\pm$ 0.17          & 0.92 $\pm$ 0.02 $\cuparrow$    & \multicolumn{1}{c|}{0.92 $\pm$ 0.02 $\cuparrow$}     & 0.01 $\pm$ 0.00        & 0.04 $\pm$ 0.02 $\cuparrow$        & 0.02 $\pm$ 0.02 $\keep$                 & 4.42    \\
\multicolumn{1}{c|}{\tt vegetable soil}      & 0.40 $\pm$ 0.10          & 0.19 $\pm$ 0.04 $\cdownarrow$  & \multicolumn{1}{c|}{0.18 $\pm$ 0.04 $\cdownarrow$}   & 0.35 $\pm$ 0.01        & 0.35 $\pm$ 0.01 $\keep$                   & 0.35 $\pm$ 0.00 $\keep$                 & 5.44   \\
\multicolumn{1}{c|}{\tt cotton crop}         & 0.96 $\pm$ 0.06          & 0.80 $\pm$ 0.06 $\cdownarrow$  & \multicolumn{1}{c|}{0.70 $\pm$ 0.08 $\cdownarrow$}   & 0.89 $\pm$ 0.02        & 0.90 $\pm$ 0.01 $\keep$                   & 0.90 $\pm$ 0.01 $\keep$                 & 11.46  \\ \bottomrule
\end{tabular}%
}
\caption{The model TPR under scenario 1, StatLog. The normal class \texttt{very damp grey soil} is similar to the abnormal training class \texttt{damp grey soil}. Their $L^2$ distance = 3.63.}
\label{tab:satlog-recall-case1}
\end{table*}

\begin{table*}[ht]
\centering
{\small training normal = \texttt{normal}, training abnormal = \texttt{NB-10ms}} \\
\resizebox{0.95\textwidth}{!}{%
\begin{tabular}{@{}cccc|ccc |c@{}}
\toprule
\multicolumn{1}{c|}{{Test data}} & \textbf{Deep SVDD} & \textbf{Deep SAD} & \textbf{HSC}  & \textbf{AE} & \textbf{SAE} & \textbf{ABC} & $L^2$  to \texttt{NB-10ms}\\ \hline
\multicolumn{1}{c|}{{\tt NB-10ms}} & 0.03 $\pm$ 0.01   & 0.99 $\pm$ 0.02 $\cuparrow$   & \multicolumn{1}{c|}{0.99 $\pm$ 0.01 $\cuparrow$}    & 0.02 $\pm$ 0.01   & 0.97 $\pm$ 0.05 $\cuparrow$    & 0.92 $\pm$ 0.03 $\cuparrow$  & 0         \\ \hline
\multicolumn{1}{c|}{\tt NB-5ms}    & 0.02 $\pm$ 0.00   & 0.93 $\pm$ 0.05 $\cuparrow$   & \multicolumn{1}{c|}{0.82 $\pm$ 0.07 $\cuparrow$}    & 0.00 $\pm$ 0.00   & 0.96 $\pm$ 0.02 $\cuparrow$    & 0.99 $\pm$ 0.01 $\cuparrow$  & 6.22       \\
\multicolumn{1}{c|}{\tt WB-nlos}        & 0.99 $\pm$ 0.00   & 0.38 $\pm$ 0.08 $\cdownarrow$ & \multicolumn{1}{c|}{0.43 $\pm$ 0.10 $\cdownarrow$}  & 0.89 $\pm$ 0.03   & 0.54 $\pm$ 0.02 $\cdownarrow$  & 0.47 $\pm$ 0.02 $\cdownarrow$  & 32.40    \\
\multicolumn{1}{c|}{\tt WB-los}         & 0.99 $\pm$ 0.00   & 0.50 $\pm$ 0.10 $\cdownarrow$ & \multicolumn{1}{c|}{0.53 $\pm$ 0.15 $\cdownarrow$}  & 0.92 $\pm$ 0.03   & 0.51 $\pm$ 0.07 $\cdownarrow$  & 0.50 $\pm$ 0.03 $\cdownarrow$  & 43.01  \\ \bottomrule
\end{tabular}%
}
\caption{The model TPR under scenario 1, Cellular Spectrum Misuse. The normal class is similar to the abnormal training class \texttt{NB-10ms}, and the $L^2$ distance between the two is 6.17.}
\label{tab:imagenet-recall-case1}
\end{table*}


\para{\bf Scenario 2.} We consider scenario 2 where the training normal set is visually dissimilar to the training abnormal set. The detailed TPR result of all six models, and three real-life datasets are in Table~\ref{tab:fmnist-recall-case2} --~\ref{tab:imagenet-recall-case2}. Like the above, in each table, the abnormal test classes are sorted by decreasing similarity to the abnormal training class. Like the above, $\cuparrow$ indicates that the supervised model has a higher recall than the semi-supervised model;  $\cdownarrow$ indicates the other direction.

Different from Scenario 1, here we observe mostly upward changes. Again we observe poorer performance of AE, SAE, ABC on  StatLog  compared to the hypersphere-based models.

\begin{table*}[ht]
\centering
{\small training normal = \texttt{top}, training abnormal = \texttt{sneaker}}
\resizebox{0.99\textwidth}{!}{%
\begin{tabular}{@{}cccc|ccc|c@{}}
\toprule
\multicolumn{1}{c|}{{Test data}} & \textbf{Deep SVDD} & \textbf{Deep SAD} & \textbf{HSC}  & \textbf{AE} & \textbf{SAE} & \textbf{ABC} & $L^2$ to \texttt{sneaker} \\ \hline
\multicolumn{1}{c|}{{\tt sneaker}}  &0.55 $\pm$ 0.09    &1.00 $\pm$ 0.00 $\cuparrow$   & \multicolumn{1}{c|}{1.00 $\pm$ 0.00 $\cuparrow$}    & 0.74 $\pm$ 0.09    &1.00 $\pm$ 0.00 $\cuparrow$    &1.00 $\pm$ 0.00 $\cuparrow$    &0 \\ \hline
\multicolumn{1}{c|}{\tt sandal}   & 0.30 $\pm$ 0.04     &0.99 $\pm$ 0.01 $\cuparrow$   & \multicolumn{1}{c|}{0.98 $\pm$ 0.02 $\cuparrow$}    & 0.82 $\pm$ 0.02   & 1.00 $\pm$ 0.00 $\cuparrow$     & 1.00 $\pm$ 0.00 $\cuparrow$            &0.02   \\
\multicolumn{1}{c|}{\tt boot}     & 0.92 $\pm$ 0.03     &1.00 $\pm$  0.00 $\cuparrow$  & \multicolumn{1}{c|}{0.97 $\pm$ 0.02 $\cuparrow$}    & 0.98 $\pm$ 0.02   & 1.00 $\pm$ 0.00 $\cuparrow$         & 1.00 $\pm$ 0.00 $\cuparrow$            &0.07 \\
\multicolumn{1}{c|}{\tt bag}      & 0.49 $\pm$ 0.07     &0.80 $\pm$ 0.05 $\cuparrow$   & \multicolumn{1}{c|}{0.81 $\pm$ 0.11 $\cuparrow$}    & 0.70 $\pm$ 0.03    & 0.84 $\pm$ 0.03 $\cuparrow$         & 0.82 $\pm$ 0.03 $\cuparrow$             &0.07 \\
\multicolumn{1}{c|}{\tt shirt}    & 0.09 $\pm$ 0.01      &0.11 $\pm$ 0.02 $\cuparrow$  & \multicolumn{1}{c|}{0.12 $\pm$ 0.01 $\cuparrow$}    & 0.12 $\pm$ 0.01    &0.13 $\pm$ 0.01 $\cuparrow$                   & 0.15 $\pm$ 0.01 $\cuparrow$   &0.10 \\
\multicolumn{1}{c|}{\tt trouser}  & 0.32 $\pm$ 0.09      &0.31 $\pm$ 0.10 $\keep$   & \multicolumn{1}{c|}{0.11 $\pm$ 0.12 $\cdownarrow$}  & 0.58 $\pm$ 0.04    & 0.58 $\pm$ 0.03 $\keep$   &  0.58 $\pm$ 0.05 $\keep$           &0.12 \\
\multicolumn{1}{c|}{\tt dress}    & 0.16 $\pm$ 0.03      &0.16 $\pm$ 0.04 $\keep$             & \multicolumn{1}{c|}{0.11 $\pm$ 0.01 $\cdownarrow$}  & 0.11 $\pm$ 0.01     &0.11 $\pm$ 0.01 $\keep$    & 0.12 $\pm$ 0.01 $\keep$  & 0.13 \\
\multicolumn{1}{c|}{\tt pullover} & 0.13 $\pm$ 0.02      &0.13 $\pm$ 0.03 $\keep$             & \multicolumn{1}{c|}{0.14 $\pm$ 0.05 $\keep$}    & 0.19 $\pm$ 0.02    & 0.21 $\pm$ 0.03 $\keep$   & 0.19 $\pm$ 0.02 $\keep$           &0.13 \\
\multicolumn{1}{c|}{\tt coat}     & 0.14 $\pm$ 0.03      &0.13 $\pm$ 0.03 $\keep$             & \multicolumn{1}{c|}{0.13 $\pm$ 0.06 $\keep$}   & 0.15 $\pm$ 0.02     & 0.16 $\pm$ 0.02 $\keep$    & 0.15 $\pm$ 0.02 $\keep$   &0.14 \\ \bottomrule
\end{tabular}%
}
\caption{The model TPR under scenario 2, Fashion-MNIST. The normal class {\tt top} is dissimilar to the abnormal training class  {\tt sneaker}, and the $L^2$ distance between the two is 0.13.}
\label{tab:fmnist-recall-case2}
\end{table*}

\begin{table*}[ht]
\centering
{\small training normal = \texttt{very damp grey soil}, training abnormal = \texttt{red soil}} \\
\resizebox{0.99\textwidth}{!}{%
\begin{tabular}{@{}cccc|ccc |c@{}}
\toprule
\multicolumn{1}{c|}{{Test data}} & \textbf{Deep SVDD} & \textbf{Deep SAD} & \textbf{HSC}  & \textbf{AE} & \textbf{SAE} & \textbf{ABC} & $L^2$ to \texttt{red soil}\\ \hline
\multicolumn{1}{c|}{{\tt red soil}}   & 0.69 $\pm$ 0.12  & 1.00 $\pm$ 0.00 $\cuparrow$   & \multicolumn{1}{c|}{1.00  $\pm$  0.00 $\cuparrow$}     & 0.00 $\pm$ 0.00        & 0.21 $\pm$ 0.01 $\cuparrow$         & 0.20 $\pm$ 0.00 $\cuparrow$      &0         \\ \hline
\multicolumn{1}{c|}{\tt damp grey soil}  & 0.12 $\pm$ 0.05   & 0.25 $\pm$ 0.04 $\cuparrow$  & \multicolumn{1}{c|}{0.22 $\pm$ 0.04 $\cuparrow$}     & 0.00 $\pm$ 0.00        & 0.00 $\pm$ 0.00 $\keep$         & 0.00 $\pm$ 0.00 $\keep$       &4.39       \\
\multicolumn{1}{c|}{\tt grey soil}     & 0.43 $\pm$ 0.16  & 0.68  $\pm$ 0.12 $\cuparrow$    & \multicolumn{1}{c|}{0.53 $\pm$ 0.09 $\cuparrow$}     & 0.01 $\pm$ 0.00        & 0.02 $\pm$ 0.00 $\keep$        & 0.01 $\pm$ 0.00 $\keep$         &5.48    \\
\multicolumn{1}{c|}{\tt vegetable soil}      & 0.40 $\pm$ 0.10     & 0.76 $\pm$ 0.07 $\cuparrow$    & \multicolumn{1}{c|}{0.76  $\pm$ 0.07 $\cuparrow$}     & 0.35 $\pm$ 0.00        & 0.42 $\pm$ 0.02 $\cuparrow$        & 0.42 $\pm$ 0.02 $\cuparrow$           & 6.63   \\
\multicolumn{1}{c|}{\tt cotton crop}        & 0.96 $\pm$ 0.06                              & 1.00 $\pm$ 0.00 $\cuparrow$ &
\multicolumn{1}{c|}{1.00 $\pm$ 0.00 $\cuparrow$}               & 0.89 $\pm$ 0.00        & 0.93 $\pm$ 0.01 $\cuparrow$                 & 0.93  $\pm$ 0.01 $\cuparrow$  & 8.97  \\ \bottomrule
\end{tabular}%
}
\caption{The model TPR  under scenario 2, StatLog. The normal class \texttt{very damp grey soil} is dissimilar to the abnormal training class \texttt{red soil}, and the $L^2$ distance between the two is 8.48.}
\label{tab:satlog-recall-case2}
\end{table*}

\begin{table*}[ht]
\centering
{\small training normal = \texttt{normal}, training abnormal = \texttt{WB-los}} \\
\resizebox{0.99\textwidth}{!}{%
\begin{tabular}{@{}cccc|ccc |c@{}}
\toprule
\multicolumn{1}{c|}{{Test data}} & \textbf{Deep SVDD} & \textbf{Deep SAD} & \textbf{HSC}  & \textbf{AE} & \textbf{SAE} & \textbf{ABC} & $L^2$ to \texttt{WB-los}\\ \hline
\multicolumn{1}{c|}{{\tt WB-los}}     & 0.99 $\pm$ 0.00   & 1.00 $\pm$ 0.01 $\cuparrow$   & \multicolumn{1}{c|}{1.00 $\pm$ 0.01 $\cuparrow$}    & 0.92 $\pm$ 0.03   & 0.95 $\pm$ 0.04 $\cuparrow$    & 1.00 $\pm$ 0.02 $\cuparrow$  &  0        \\ \hline
\multicolumn{1}{c|}{\tt WB-nlos}      & 0.99 $\pm$ 0.00   & 1.00 $\pm$ 0.01 $\cuparrow$   & \multicolumn{1}{c|}{1.00 $\pm$ 0.00 $\cuparrow$}    & 0.89 $\pm$ 0.03   & 0.94 $\pm$ 0.03 $\cuparrow$    & 0.96 $\pm$ 0.02 $\cuparrow$  &  14.39      \\
\multicolumn{1}{c|}{\tt NB-10ms} & 0.03 $\pm$ 0.01   & 0.06 $\pm$ 0.01 $\cuparrow$   & \multicolumn{1}{c|}{0.05 $\pm$ 0.02 $\keep$}    & 0.02 $\pm$ 0.01   & 0.03 $\pm$ 0.00 $\cuparrow$    & 0.04 $\pm$ 0.01 $\keep$  & 43.01    \\
\multicolumn{1}{c|}{\tt NB-5ms}  & 0.02 $\pm$ 0.00   & 0.03 $\pm$ 0.01 $\keep$               & \multicolumn{1}{c|}{0.02 $\pm$ 0.00 $\keep$}                & 0.00 $\pm$ 0.00   & 0.02 $\pm$ 0.00 $\cuparrow$    & 0.02 $\pm$ 0.01 $\cuparrow$  & 44.37  \\ \bottomrule
\end{tabular}%
}
\caption{The model TPR under scenario 2, Cellular Spectrum Misuse. The normal class is dissimilar to the abnormal training class \texttt{WB-los}, and the $L^2$ distance between the two is 43.84.}
\label{tab:imagenet-recall-case2}
\end{table*}
\newpage

\label{sec:addconfig}

\para{\bf Scenario 3.} We run three configurations of grouped abnormal training on Fashion-MNIST (training normal: {\tt top}; training abnormal: {\tt shirt} \& {\tt sneaker}) by varying the weights of the two abnormal classes in training (0.5/0.5, 0.9/0.1, 0.1/0.9). Again $\cuparrow$ indicates that the supervised model has a higher recall than the semi-supervised model;  $\cdownarrow$ indicates the other direction. Under these settings, we observe downward bias ($\cdownarrow$) for one abnormal test class {\tt trouser} and upward bias for most other classes.

\begin{table*}[ht]
\centering
{\small training normal = \texttt{top}, training abnormal = 50\% \texttt{shirt} and 50\% \texttt{sneaker}} \\
\resizebox{1\textwidth}{!}{%
\begin{tabular}{@{}cccc|ccc |cc@{}}
\toprule
\multicolumn{1}{c|}{{Test data}} & \textbf{Deep SVDD} & \textbf{Deep SAD} & \textbf{HSC}  & \textbf{AE} & \textbf{SAE} & \textbf{ABC} & $L^2$ to \texttt{shirt} & $L^2$ to \texttt{sneaker} \\ \hline
\multicolumn{1}{c|}{{\tt shirt}}    & 0.09 $\pm$ 0.01     & 0.69 $\pm$ 0.01 $\cuparrow$    & \multicolumn{1}{c|}{0.69 $\pm$ 0.02 $\cuparrow$}     & 0.12 $\pm$ 0.01        & 0.67 $\pm$ 0.01 $\cuparrow$         & 0.66 $\pm$ 0.01 $\cuparrow$      &0 &0.10         \\
\multicolumn{1}{c|}{\tt sneaker}    & 0.55 $\pm$ 0.09     & 1.00 $\pm$ 0.00 $\cuparrow$  & \multicolumn{1}{c|}{1.00 $\pm$ 0.00 $\cuparrow$}   & 0.74 $\pm$ 0.09        & 1.00 $\pm$ 0.00 $\cuparrow$       & 1.00 $\pm$ 0.00 $\cuparrow$          &0.10 & 0    \\ \hline
\multicolumn{1}{c|}{\tt pullover}   & 0.13 $\pm$ 0.02     & 0.90 $\pm$ 0.01 $\cuparrow$    & \multicolumn{1}{c|}{0.90 $\pm$ 0.01 $\cuparrow$}     & 0.19 $\pm$ 0.02        & 0.82 $\pm$ 0.02 $\cuparrow$         & 0.83 $\pm$ 0.02 $\cuparrow$        &0.01 & 0.13       \\
\multicolumn{1}{c|}{\tt coat}       & 0.14 $\pm$ 0.03     & 0.91 $\pm$ 0.02 $\cuparrow$    & \multicolumn{1}{c|}{0.90 $\pm$ 0.01 $\cuparrow$}      & 0.15 $\pm$ 0.02        & 0.86 $\pm$ 0.02 $\cuparrow$          & 0.87 $\pm$ 0.02 $\cuparrow$            &0.01 & 0.14    \\
\multicolumn{1}{c|}{\tt dress}      & 0.17 $\pm$ 0.03     & 0.23 $\pm$ 0.04 $\cuparrow$    & \multicolumn{1}{c|}{0.24 $\pm$ 0.04 $\cuparrow$}     & 0.11 $\pm$ 0.01        & 0.19 $\pm$ 0.03 $\cuparrow$         & 0.18 $\pm$ 0.02 $\cuparrow$            &0.04 & 0.13   \\
\multicolumn{1}{c|}{\tt bag}        & 0.49 $\pm$ 0.07     & 0.63 $\pm$ 0.06 $\cuparrow$  & \multicolumn{1}{c|}{0.62 $\pm$ 0.07 $\cuparrow$}   & 0.70 $\pm$ 0.03        & 0.76 $\pm$ 0.05 $\cuparrow$       & 0.78 $\pm$ 0.03 $\cuparrow$          &0.04 & 0.07  \\
\multicolumn{1}{c|}{\tt trouser}    & 0.32 $\pm$ 0.10     & 0.05 $\pm$ 0.04 $\cdownarrow$  & \multicolumn{1}{c|}{0.04 $\pm$ 0.02 $\cdownarrow$}   & 0.59 $\pm$ 0.04        & 0.22 $\pm$ 0.08 $\cdownarrow$   & 0.34 $\pm$ 0.06 $\cdownarrow$          &0.06 & 0.12\\
\multicolumn{1}{c|}{\tt boot}       & 0.92 $\pm$ 0.03     & 0.95 $\pm$ 0.03 $\keep$ & \multicolumn{1}{c|}{0.95 $\pm$ 0.03 $\keep$}   & 0.98 $\pm$ 0.02        & 1.00 $\pm$ 0.00 $\cuparrow$       & 1.00 $\pm$ 0.00 $\cuparrow$          &0.08 & 0.07   \\
\multicolumn{1}{c|}{\tt sandal}     & 0.30 $\pm$ 0.04     & 0.92 $\pm$ 0.04 $\cuparrow$    & \multicolumn{1}{c|}{0.92 $\pm$ 0.04 $\cuparrow$}   & 0.82 $\pm$ 0.02        & 0.96 $\pm$ 0.01 $\cuparrow$       & 0.97 $\pm$ 0.01 $\cuparrow$          &0.09 & 0.02     \\ \bottomrule
\end{tabular}%
}
\caption{The model TPR under configuration 1 of weighted mixture training on Fashion-MNIST.}
\label{tab:fmnist-config-1}
\end{table*}

\begin{table*}[ht]
\centering
{\small training normal = \texttt{top}, training abnormal = 90\% \texttt{shirt} and 10\% \texttt{sneaker}} \\
\resizebox{1\textwidth}{!}{%
\begin{tabular}{@{}cccc|ccc |cc@{}}
\toprule
\multicolumn{1}{c|}{{Test data}} & \textbf{Deep SVDD} & \textbf{Deep SAD} & \textbf{HSC}  & \textbf{AE} & \textbf{SAE} & \textbf{ABC} & $L^2$ to \texttt{shirt} & $L^2$ to \texttt{sneaker} \\ \hline
\multicolumn{1}{c|}{{\tt shirt}}    & 0.09 $\pm$ 0.01     & 0.70 $\pm$ 0.01 $\cuparrow$    & \multicolumn{1}{c|}{0.70 $\pm$ 0.01 $\cuparrow$}     & 0.12 $\pm$ 0.01        & 0.72 $\pm$ 0.01 $\cuparrow$       & 0.71 $\pm$ 0.01 $\cuparrow$      &0 &0.10         \\
\multicolumn{1}{c|}{\tt sneaker}    & 0.55 $\pm$ 0.09     & 1.00 $\pm$ 0.00 $\cuparrow$    & \multicolumn{1}{c|}{1.00 $\pm$ 0.00 $\cuparrow$}   & 0.74 $\pm$ 0.09          & 1.00 $\pm$ 0.00 $\cuparrow$       & 1.00 $\pm$ 0.00 $\cuparrow$          &0.10 & 0    \\ \hline
\multicolumn{1}{c|}{\tt pullover}   & 0.13 $\pm$ 0.02     & 0.90 $\pm$ 0.01 $\cuparrow$    & \multicolumn{1}{c|}{0.89 $\pm$ 0.01 $\cuparrow$}     & 0.19 $\pm$ 0.02        & 0.84 $\pm$ 0.02 $\cuparrow$       & 0.84 $\pm$ 0.02 $\cuparrow$        &0.01 & 0.13       \\
\multicolumn{1}{c|}{\tt coat}       & 0.14 $\pm$ 0.03     & 0.91 $\pm$ 0.02 $\cuparrow$    & \multicolumn{1}{c|}{0.91 $\pm$ 0.02 $\cuparrow$}      & 0.15 $\pm$ 0.02        & 0.91 $\pm$ 0.02 $\cuparrow$        & 0.90 $\pm$ 0.02 $\cuparrow$            &0.01 & 0.14    \\
\multicolumn{1}{c|}{\tt dress}      & 0.17 $\pm$ 0.03     & 0.23 $\pm$ 0.03 $\cuparrow$    & \multicolumn{1}{c|}{0.24 $\pm$ 0.03 $\cuparrow$}     & 0.11 $\pm$ 0.01        & 0.19 $\pm$ 0.03 $\cuparrow$       & 0.20 $\pm$ 0.03 $\cuparrow$            &0.04 & 0.13   \\
\multicolumn{1}{c|}{\tt bag}        & 0.49 $\pm$ 0.07     & 0.56 $\pm$ 0.08 $\keep$    & \multicolumn{1}{c|}{0.57 $\pm$ 0.07 $\keep$}     & 0.70 $\pm$ 0.03        & 0.67 $\pm$ 0.06 $\keep$     & 0.68 $\pm$ 0.05 $\keep$          &0.04 & 0.07  \\
\multicolumn{1}{c|}{\tt trouser}    & 0.32 $\pm$ 0.10     & 0.06 $\pm$ 0.04 $\cdownarrow$  & \multicolumn{1}{c|}{0.06 $\pm$ 0.03 $\cdownarrow$}   & 0.59 $\pm$ 0.04        & 0.10 $\pm$ 0.06 $\cdownarrow$     & 0.20 $\pm$ 0.08 $\cdownarrow$          &0.06 & 0.12\\
\multicolumn{1}{c|}{\tt boot}       & 0.92 $\pm$ 0.03     & 0.87 $\pm$ 0.08 $\keep$   & \multicolumn{1}{c|}{0.88 $\pm$ 0.05 $\keep$}   & 0.98 $\pm$ 0.02        & 0.99 $\pm$ 0.01 $\keep$        & 0.99 $\pm$ 0.00 $\keep$          &0.08 & 0.07   \\
\multicolumn{1}{c|}{\tt sandal}     & 0.30 $\pm$ 0.04     & 0.84 $\pm$ 0.06 $\cuparrow$    & \multicolumn{1}{c|}{0.83 $\pm$ 0.05 $\cuparrow$}   & 0.82 $\pm$ 0.02          & 0.90 $\pm$ 0.02 $\cuparrow$       & 0.94 $\pm$ 0.02 $\cuparrow$          &0.09 & 0.02     \\ \bottomrule
\end{tabular}%
}
\caption{The model TPR under configuration 2 of weighted mixture training on Fashion-MNIST.}
\label{tab:fmnist-config-2}
\end{table*}

\begin{table*}[ht]
\centering
{\small training normal = \texttt{top}, training abnormal = 10\% \texttt{shirt} and 90\% \texttt{sneaker}} \\
\resizebox{1\textwidth}{!}{%
\begin{tabular}{@{}cccc|ccc |cc@{}}
\toprule
\multicolumn{1}{c|}{{Test data}} & \textbf{Deep SVDD} & \textbf{Deep SAD} & \textbf{HSC}  & \textbf{AE} & \textbf{SAE} & \textbf{ABC} & $L^2$ to \texttt{shirt} & $L^2$ to \texttt{sneaker} \\ \hline
\multicolumn{1}{c|}{{\tt shirt}}    & 0.09 $\pm$ 0.01     & 0.61 $\pm$ 0.02 $\cuparrow$    & \multicolumn{1}{c|}{0.60 $\pm$ 0.02 $\cuparrow$}     & 0.12 $\pm$ 0.01        & 0.54 $\pm$ 0.02 $\cuparrow$       & 0.54 $\pm$ 0.01 $\cuparrow$      &0 &0.10         \\
\multicolumn{1}{c|}{\tt sneaker}    & 0.55 $\pm$ 0.09     & 1.00 $\pm$ 0.00 $\cuparrow$    & \multicolumn{1}{c|}{1.00 $\pm$ 0.00 $\cuparrow$}     & 0.74 $\pm$ 0.09        & 1.00 $\pm$ 0.00 $\cuparrow$       & 1.00 $\pm$ 0.00 $\cuparrow$          &0.10 & 0   \\ \hline
\multicolumn{1}{c|}{\tt pullover}   & 0.13 $\pm$ 0.02     & 0.85 $\pm$ 0.02 $\cuparrow$    & \multicolumn{1}{c|}{0.84 $\pm$ 0.02 $\cuparrow$}     & 0.19 $\pm$ 0.02        & 0.74 $\pm$ 0.03 $\cuparrow$       & 0.74 $\pm$ 0.03 $\cuparrow$        &0.01 & 0.13       \\
\multicolumn{1}{c|}{\tt coat}       & 0.14 $\pm$ 0.03     & 0.79 $\pm$ 0.03 $\cuparrow$    & \multicolumn{1}{c|}{0.77 $\pm$ 0.03 $\cuparrow$}     & 0.15 $\pm$ 0.02        & 0.67 $\pm$ 0.04 $\cuparrow$        & 0.68 $\pm$ 0.03 $\cuparrow$            &0.01 & 0.14    \\
\multicolumn{1}{c|}{\tt dress}      & 0.17 $\pm$ 0.03     & 0.13 $\pm$ 0.03 $\cdownarrow$  & \multicolumn{1}{c|}{0.11 $\pm$ 0.03 $\cdownarrow$}   & 0.11 $\pm$ 0.01        & 0.12 $\pm$ 0.01 $\keep$                  & 0.11 $\pm$ 0.01 $\keep$           &0.04 & 0.13   \\
\multicolumn{1}{c|}{\tt bag}        & 0.49 $\pm$ 0.07     & 0.82 $\pm$ 0.05 $\cuparrow$    & \multicolumn{1}{c|}{0.84 $\pm$ 0.05 $\cuparrow$}     & 0.70 $\pm$ 0.03        & 0.85 $\pm$ 0.03 $\cuparrow$     & 0.86 $\pm$ 0.02 $\cuparrow$          &0.04 & 0.07  \\
\multicolumn{1}{c|}{\tt trouser}    & 0.32 $\pm$ 0.10     & 0.10 $\pm$ 0.08 $\cdownarrow$  & \multicolumn{1}{c|}{0.05 $\pm$ 0.05 $\cdownarrow$}   & 0.59 $\pm$ 0.04        & 0.45 $\pm$ 0.07 $\cdownarrow$     & 0.50 $\pm$ 0.05 $\cdownarrow$          &0.06 & 0.12\\
\multicolumn{1}{c|}{\tt boot}       & 0.92 $\pm$ 0.03     & 1.00 $\pm$ 0.00 $\cuparrow$    & \multicolumn{1}{c|}{0.99 $\pm$ 0.01 $\cuparrow$}     & 0.98 $\pm$ 0.02        & 0.98 $\pm$ 0.00 $\keep$                  & 1.00 $\pm$ 0.00 $\keep$           &0.08 & 0.07   \\
\multicolumn{1}{c|}{\tt sandal}     & 0.30 $\pm$ 0.04     & 0.99 $\pm$ 0.01 $\cuparrow$  & \multicolumn{1}{c|}{0.83 $\pm$ 0.01 $\cuparrow$}     & 0.82 $\pm$ 0.02        & 0.98 $\pm$ 0.00 $\cuparrow$       & 0.99 $\pm$ 0.00 $\cuparrow$          &0.09 & 0.02     \\ \bottomrule

\end{tabular}%
}
\caption{The model TPR under configuration 3 of weighted mixture training on Fashion-MNIST.}
\label{tab:fmnist-config-3}
\end{table*}

\end{document}